\documentclass[runningheads]{llncs}

\usepackage{eccv}

\usepackage{eccvabbrv}

\usepackage{graphicx}
\usepackage{booktabs}
\usepackage{cuted}
\usepackage{capt-of}
\usepackage{multirow}
\usepackage{wrapfig}
\usepackage{bm}

\usepackage[accsupp]{axessibility}  

\usepackage[pagebackref,breaklinks,colorlinks,citecolor=eccvblue]{hyperref}

\usepackage{orcidlink}

\begin{document}

\title{Learning Local Pattern Modularization for Point Cloud Reconstruction from Unseen Classes} 

\titlerunning{Unseen Point Cloud Reconstruction with Local Pattern}




\author{Chao Chen\inst{1} \and
Yu-Shen Liu\inst{1}\thanks{The corresponding author is Yu-Shen Liu. This work was supported by National Key R$\&$D Program of China (2022YFC3800600), the National Natural Science Foundation of China (62272263, 62072268), and in part by Tsinghua-Kuaishou Institute of Future Media Data.} \and
Zhizhong Han\inst{2}}
\authorrunning{C. Chen et al.}
%
\institute{School of Software, Tsinghua University, Beijing, China \and
Department of Computer Science, Wayne State University, Detroit, USA\\
\email{chenchao19@mails.tsinghua.edu.cn, liuyushen@tsinghua.edu.cn, h312h@wayne.edu}}

\maketitle

\begin{abstract}
It is challenging to reconstruct 3D point clouds in unseen classes from single 2D images. Instead of object-centered coordinate system, current methods generalized global priors learned in seen classes to reconstruct 3D shapes from unseen classes in viewer-centered coordinate system. However, the reconstruction accuracy and interpretability are still eager to get improved. To resolve this issue, we introduce to learn local pattern modularization for reconstructing 3D shapes in unseen classes, which achieves both good generalization ability and high reconstruction accuracy. Our insight is to learn a local prior which is class-agnostic and easy to generalize in object-centered coordinate system. Specifically, the local prior is learned via a process of learning and customizing local pattern modularization in seen classes. During this process, we first learn a set of patterns in local regions, which is the basis in the object-centered coordinate system to represent an arbitrary region on shapes across different classes. Then, we modularize each region on an initially reconstructed shape using the learned local patterns. Based on that, we customize the local pattern modularization using the input image by refining the reconstruction with more details. Our method enables to reconstruct high fidelity point clouds from unseen classes in object-centered coordinate system without requiring a large number of patterns or any additional information, such as segmentation supervision or camera poses. Our experimental results under widely used benchmarks show that our method achieves the state-of-the-art reconstruction accuracy for shapes from unseen classes. The code is available at \url{https://github.com/chenchao15/Unseen}.
\end{abstract}

\begin{figure}[tb]
\centering
\includegraphics[width=\textwidth]{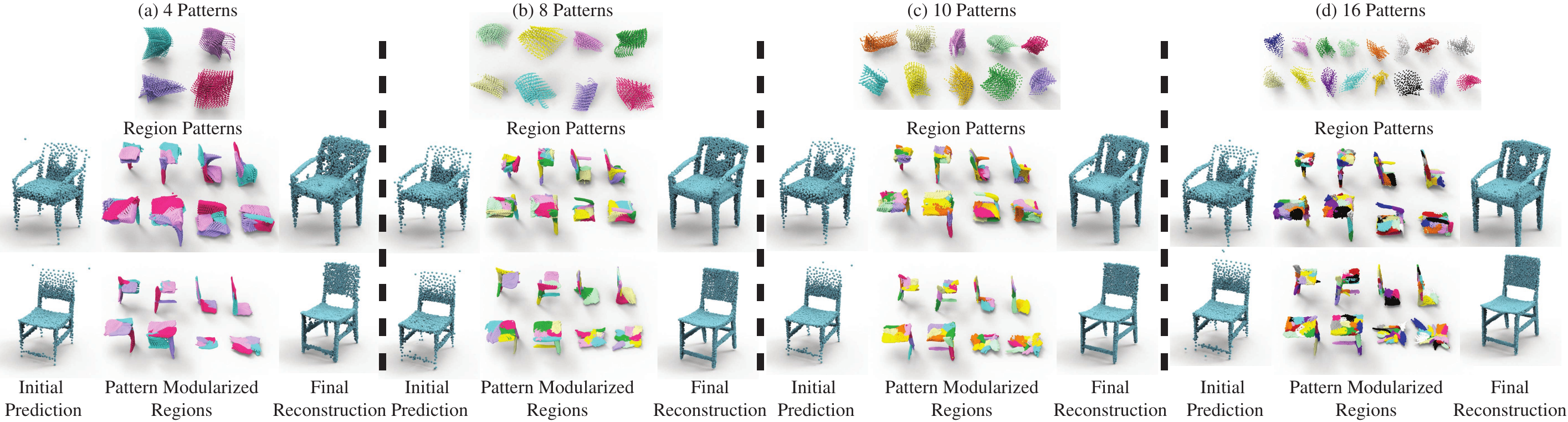}
\captionof{figure}{We learn to reconstruct shapes from unseen classes by learning a local class-agnostic prior with region patterns, such as (a) 4 patterns, (b) 8 patterns, (c) 10 patterns, or (d) 16 patterns. Using region patterns, we modularize each region of the initial reconstruction. Based on that, we obtain our final reconstruction by customizing these pattern modularized regions.
}
\end{figure}

\section{Introduction}
\label{sec:intro}
It is challenging and vital to reconstruct point clouds from unseen classes. A widely used strategy~\cite{FanSG17,DBLP:conf/cvpr/SoltaniH0KT17,DBLP:conf/iccv/NguyenCK019,Groueix_2018_CVPR,zhizhongiccv2021finepoints} for point cloud reconstruction is to learn a global prior from 2D images to 3D point clouds in object-centered coordinate system. The global prior is the key for good reconstruction accuracy, since shapes from the same classes are well aligned in the object-centered coordinate system during training. However, the global prior can not generalize well to infer shapes from unseen classes.

Some methods learn global priors in viewer-centered coordinate system which achieve better generalization to unseen classes, where the ground truth 3D shape is rotated
to match the pose of the object in the input image. These methods can reasonably generalize to unseen classes by requiring additional information to support the generalization, such as camera parameters~\cite{NIPS2018_7494,DBLP:conf/eccv/WangF20,bautista2020generalization} or category shape prior~\cite{DBLP:conf/iccv/WallaceH19}. However, they still struggle to improve reconstruction accuracy for unseen classes due to the large geometry variation across classes and various camera poses. More recent methods~\cite{bautista2020generalization,3DVUnseen,Bechtold2021HPN,Xian2022gin,Zhizhong2021icml,predictivecontextpriors2022,onsurfacepriors2022,BaoruiNoise2NoiseMapping} represent 3D shapes as implicit functions to increase reconstruction accuracy. However, these methods require a large number of queries as training samples for each shape, and lack of interpretability due to the limited capability of implicit functions for representing open surfaces of parts.

To improve accuracy and interpretability, we propose to learn local pattern modularization for reconstructing 3D shapes from unseen classes. Our insight is to learn a local prior in object-centered coordinate system which can not only generalize well to unseen classes but also reconstruct high fidelity point clouds. Our idea comes from the observation that shapes from different classes may share some similar local structures, which makes it feasible to learn a local prior that is class-agnostic. Therefore, rather than a global prior, we learn a local prior from 3D local regions which can be generalized better to unseen classes.
Specifically, we aim to learn a set of patterns in local regions, as demonstrated in Fig.~\ref{fig:pattern}, which can be used as the basis in the object-centered coordinate system to represent each region on shapes across different classes. From an input image, we first predict an initial shape reconstruction, and then modularize each region using the learned region patterns.
Based on that, we learn to customize these pattern modularized regions according to the input image by refining them with more details using a learned modularization shift. Our method enables to reconstruct high fidelity point clouds from unseen classes in object-centered coordinate system without requiring a large number of patterns or any additional information, such as segmentation supervision or camera poses. Our experimental results under widely used benchmarks show that our method achieves the state-of-the-art reconstruction accuracy for shapes from unseen classes. Our contributions are listed below.

\begin{itemize}
\item We introduce to learn local pattern modularization for 3D shape reconstruction from unseen classes. By further customizing the modularization, we obtain a local prior which gets better generalized to unseen classes than current global priors.
\item We justify the feasibility of class-agnostic local prior in the object-centered coordinate system, which significantly improves the reconstruction accuracy in point cloud reconstruction from unseen classes.
\item Our method achieves the state-of-the-art reconstruction accuracy of point clouds in both seen and unseen classes under the widely used benchmarks.
\end{itemize}

\section{Related Work}
Deep learning based 3D shape reconstruction has made a big progress with different 3D representations including voxel grids~\cite{YanNIPS2016,TulsianiZEM17,ChoyXGCS16}, triangle meshes~\cite{liu2019softras,Zhizhong2016,Zhizhong2016b,chen2019dibrender}, point clouds~\cite{Han2019ShapeCaptionerGCacmmm,cvprpoint2017,InsafutdinovD18,p2seq18,l2g2019,Groueix_2018_CVPR,handrwr2020,MAPVAE19,bednarik2020,pmpnet++}, and implicit functions~\cite{MeschederNetworks,Park_2019_CVPR,seqxy2seqzeccv2020paper,chen2018implicit_decoder,Jiang2019SDFDiffDRcvpr,mildenhall2020nerf,ben2021digs,10506631,Zhou2022CAP-UDF,chaompi2022,feng2022np,NeuralTPS,chao2023gridpull}. The widely used strategy aims to leverage deep learning models to learn a global prior for shape reconstruction from 2D images. In the following, we focus on reviewing studies for point clouds reconstruction.

\noindent\textbf{Supervised Point Clouds Reconstruction from Seen Classes. }PointNet~\cite{cvprpoint2017} is a pioneer work in point clouds understanding. Supervised methods learn to reconstruct point clouds from pairs of 2D image and its corresponding point cloud. With an encoder for 2D image understanding, Fan et al.~\cite{FanSG17} built up an encoder-decoder architecture with various shortcuts to reconstruct the point clouds. By fusing multiple depth and silhouette images generated from different view angles, Soltani et al.~\cite{DBLP:conf/cvpr/SoltaniH0KT17} reconstruct dense point clouds using a 2D neural network. Nguyen et al~\cite{DBLP:conf/iccv/NguyenCK019} tried to deform a random point cloud to the object shape with image feature blending to increase the point cloud reconstruction accuracy. Toward dense point cloud reconstruction with texture, Hu et al.~\cite{Hu2019LearningTG} reformulated the reconstruction as an object coordinate map prediction and shape completion problem. AtlasNet~\cite{Groueix_2018_CVPR} generates a point cloud as multiple 3D patches which are transformed from a set of 2D sampled points.

\noindent\textbf{Unsupervised Point Clouds Reconstruction from Seen Classes. }Without ground truth point clouds as supervision, unsupervised methods learn to reconstruct point clouds using various differentiable renders to compare the reconstructed point clouds and ground truth 2D images. Lin et al.~\cite{lin2018learning} introduced a pseudo-renderer to model the visibility using pooling in the dense points projection. Other rendering based methods~\cite{InsafutdinovD18,Navaneet2019,navaneet2019differ,Yifan:DSS:2019a} leveraged surface splatting~\cite{Yifan:DSS:2019a}, Gaussian functions in 3D space~\cite{InsafutdinovD18} or on 2D images~\cite{Navaneet2019,navaneet2019differ} to rasterize point clouds. CapNet~\cite{Navaneet2019} introduced a loss to match rendered pixels and pixels on ground truth silhouette images. Without pixel-wise interpolation, visibility handling, or shading in rendering, DRWR~\cite{handrwr2020} directly inferred losses to adjust each 3D point from pixel values and its 2D projection. 

\noindent\textbf{3D Shape Reconstruction from Unseen Classes. }The studies mentioned above only learn a global prior for the reconstruction of point clouds from classes that have been seen in the training. However, these learned prior is hard to be generalized to reconstruct point clouds from unseen classes. To learn more generalized global prior, GenRe~\cite{NIPS2018_7494} disentangled geometric projections from shape reconstruction, where depth prediction and spherical map inpainting are used for class-agnostic reconstruction. With a provided category shape prior, Wallace et al.~\cite{DBLP:conf/iccv/WallaceH19} introduced few-shot 3D shape generation by category agnostic refinement of the provided category-specific prior. Similarly, GSIR~\cite{DBLP:conf/eccv/WangF20} jointly learned interpretation and reconstruction to capture class-agnostic prior to recover 3D structures as cuboids. Recent work~\cite{bautista2020generalization,3DVUnseen,Bechtold2021HPN,Xian2022gin} employed implicit functions for 3D reconstruction from unseen classes. These methods extend the potentials of generalization for unseen classes shown in some local implicit function based methods~\cite{Genova_2020_CVPR,jiang2020lig,Peng2020ECCV,DBLP:conf/eccv/ChabraLISSLN20}. However, these methods require a large number of queries as training samples for each shape, and lack of interpretability due to the incapability of implicit functions for representing open surfaces of parts. Instead, we use point clouds to interpret the reconstruction with much fewer points for each shape. Recent large visual model~\cite{nichol2022point,jun2023shap} aims to learn reconstruction on a large scale of classes.

Different from these methods, our method learns a local prior for point clouds reconstruction without requiring camera parameters or category shape prior, which is much more generalizable to unseen classes. Moreover, the customization of pattern modularized regions also enables us to reconstruct point clouds in object-centered coordinate system, which achieves much higher accuracy.

\begin{figure}[tb]
  \centering
   \includegraphics[width=0.95\linewidth]{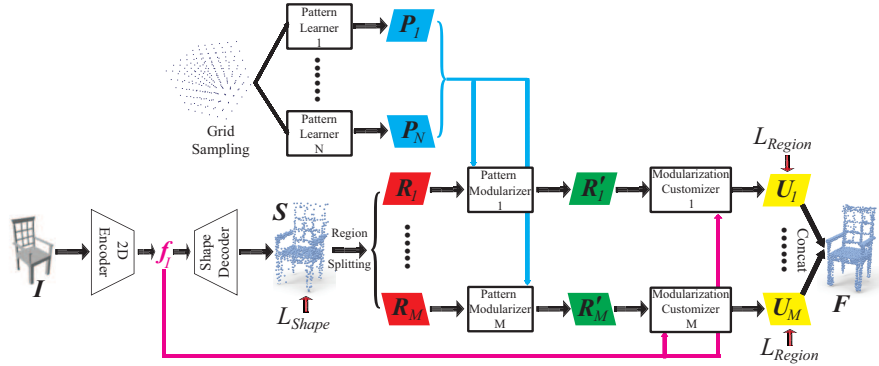}
  %
  %
\caption{\label{fig:Overview}The demonstration of our method. We aim to reconstruct a point clouds $\bm{F}$ from input image $\bm{I}$, where $\bm{F}$ may come from classes that are not seen during training.
}
\end{figure}

\section{Method}
\label{sec:method}
\noindent\textbf{Overview. }Our framework is demonstrated in Fig.~\ref{fig:Overview}. We aim to reconstruct a point cloud $\bm{F}$ from an input image $\bm{I}$, where $\bm{F}$ is from a class that is not seen during training. We represent point clouds involved in our network in the object-centered coordinate system.

We first reconstruct an initial shape prediction $\bm{S}$ from image $\bm{I}$ using an encoder and decoder network. The 2D encoder extracts the information of $\bm{I}$ as a latent code $\bm{f}_I$, which is further used to generate the initial shape prediction $\bm{S}$ by a shape decoder. Here, we leverage a shape constraint $L_{Shape}$ to make the predicted $\bm{S}$ plausible.

Then, we split the initial shape prediction $\bm{S}$ into regions $\{\bm{R}_m, m\in[1,M]\}$ to reduce the bias on seen classes during training, since regions across different classes may share similar local structures. $\{\bm{R}_m\}$ is further used to learn the region patterns $\{\bm{P}_n, n\in[1,N]\}$. We use $\{\bm{P}_n\}$ as the basis to represent various local regions across different classes in the object-centered coordinate system, which is one key to improve the generalization ability. We learn each region pattern $\bm{P}_n$ by transforming a grid sampling using a pattern learner. We use all region patterns $\{\bm{P}_n\}$ to modularize each region $\bm{R}_m$ in a pattern modularizer, so that each region can be represented based on the same set of patterns $\{\bm{P}_n\}$, which results in a pattern modularized region $\bm{R}_m'$. Learning local pattern modularization is our first step to learn a local prior for unseen classes.

We further learn to customize each pattern modularized region $\bm{R}_m'$ in a modularization customizer. Since $\bm{R}_m'$ only represents the structure of regions but without geometry details, we introduce to leverage the input image to provide geometry details, which is another key to improve the generalization ability. Our insight here is that getting images involved in part generation would further achieve class-agnostic reconstruction. The modularization customizer customizes $\bm{R}_m'$ into a pattern customized region $\bm{U}_m$ according to the latent code $\bm{f}_I$ of input image $\bm{I}$. This aims to push the modularization customizer to generate regions $\bm{U}_m$ that fits $\bm{f}_I$ better without a bias on classes. We push the modularization customizer to produce a set of pattern customized regions $\{\bm{U}_m, m\in[1,M]\}$ which form the final shape reconstruction $\bm{F}$ by concatenation. We further add a region constraint $L_{Region}$ to $\{\bm{U}_m\}$ to supervise the customization procedure.

Finally, we train our network to capture a local prior by minimizing a loss function combining $L_{Shape}$ and $L_{Region}$,

\begin{equation}
\label{eq:loss}
\begin{aligned}
L=L_{Region}+\alpha L_{Shape},
\end{aligned}
\end{equation}

\noindent where $\alpha$ is a balance weight and we will elaborate on $L_{Shape}$ and $L_{Region}$ in the following.

\noindent\textbf{Initial Shape Prediction. }We start from learning a mapping from input image $\bm{I}$ to a shape $\bm{S}\in\mathbb R^{S\times 3}$. The mapping produces an intermediate representation as a latent code $\bm{f}_I\in\mathbb{R}^{1\times H}$ to bridge the image and shape space. We aim to capture a weak global prior to make the initial shape prediction $\bm{S}$ plausible, which helps our network to have a good start without relying on specific classes. We leverage a Chamfer Distance (CD) to generate a plausible $\bm{S}$ below,

\begin{equation}
\label{eq:cd}
\begin{aligned}
L_{Shape}=\sum_{g\in\bm{G}}\min_{s\in\bm{S}}||s-g||_2+\sum_{s\in\bm{S}}\min_{g\in\bm{G}}||s-g||_2,
\end{aligned}
\end{equation}

\noindent where $\bm{G}\in\mathbb{R}^{G\times 3}$ is the ground truth point clouds and we leverage a small weight $\alpha=0.1$ in front of $L_{Shape}$ in Eq.~\ref{eq:loss} to keep the global prior weak and not biased on seen classes during training.

\noindent\textbf{Region Splitting. }We split initial shape prediction $\bm{S}$ into regions $\{\bm{R}_m\in\mathbb R^{R\times 3}, m\in[1,M]\}$ to learn the region patterns in the object-centered coordinate system. During training, we determine the range of each region $\bm{R}_m$ by voxelizing the bounding box of ground truth $\bm{G}$, such that $\bm{G}=\{\bm{G}_m, m\in[1,M]\}$. We split each edge of the bounding box into $M^{1/3}$ segments,
and regard the points on $\bm{S}$ which are located in the same voxel of $\bm{G}_m$ as $\bm{R}_m$.
While we get $\bm{R}_m$ during test by directly voxelizing the bounding box of the initial shape prediction $\bm{S}$. Note that we keep the number of points in each region $\bm{R}_m$ the same by padding zero points for more convenient manipulation in network.

\begin{wrapfigure}[11]{r}{0.55\linewidth}
\includegraphics[width=\linewidth]{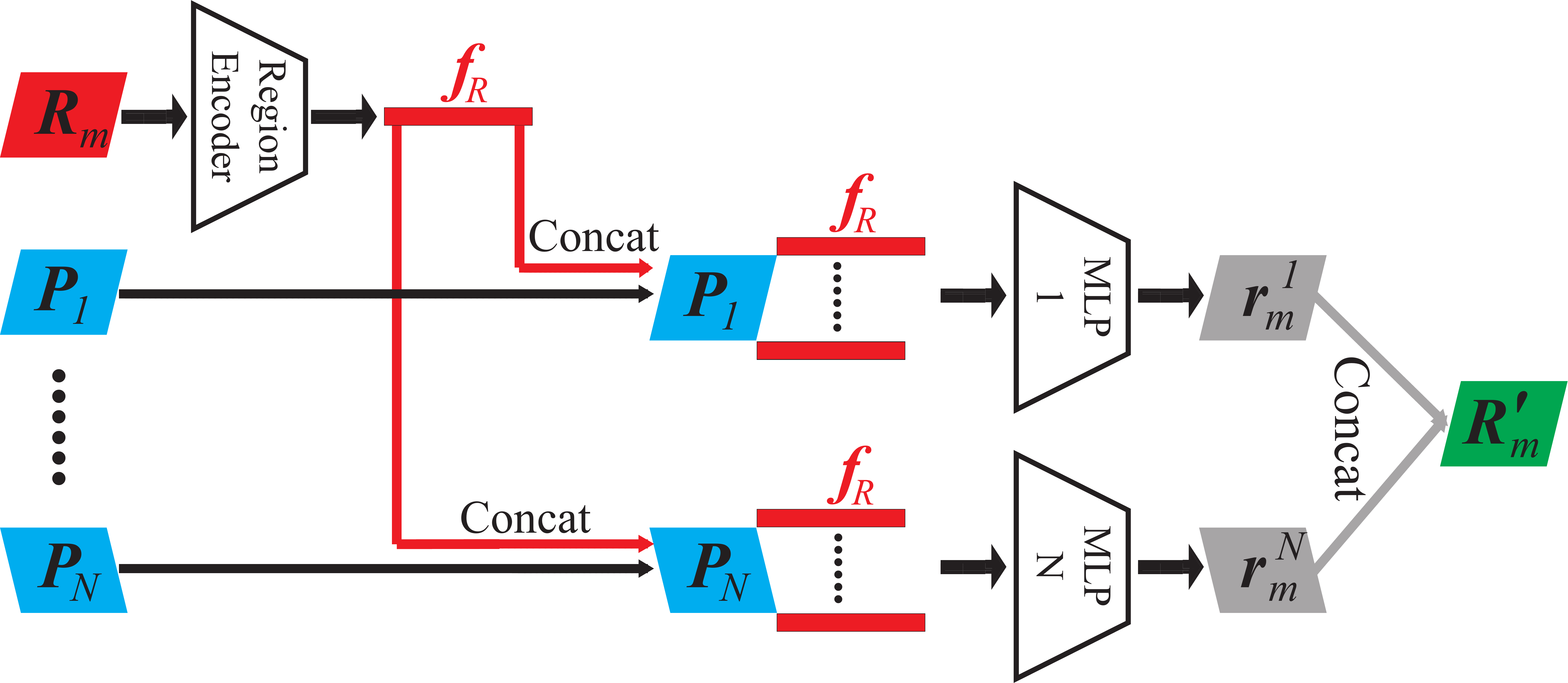}
\caption{\label{fig:mod}The architecture of pattern modularizer.}
\end{wrapfigure}

\noindent\textbf{Pattern Learner. }We learn region patterns $\{\bm{P}_n\in\mathbb R^{P\times 3}, n\in[1,N]\}$ in local region coordinate system by translating each region to the origin, as defined in Eq.~\ref{eq:c2d}. This centering makes structures in local regions comparable to each other, which also helps region patterns easily learn more reasonable common structures.

\begin{equation}
\label{eq:c2d}
\begin{aligned}
\bm{R}_m\gets\bm{R}_m-\bm{c}_m, \text{where} \, \bm{c}_m=mean(\bm{R}_m).
\end{aligned}
\end{equation}

We use $N$ pattern learners to learn $\{\bm{P}_n\}$. Each learner transforms a set of points sampled on the voxel grid into a region pattern. We share the similar idea of AtlasNet~\cite{DBLP:conf/nips/DeprelleGFKRA19,Groueix_2018_CVPR} to generate a pattern with strong neighboring relationship. However, we translate the same set of sampled points to make each pattern learner have a different start, which results in more discriminative region patterns. All region patterns $\{\bm{P}_n\}$ are involved in the following pattern modularization process.


\noindent\textbf{Pattern Modularizer. }We modularize each region $\bm{R}_m$ using all the region patterns $\{\bm{P}_n\}$ in pattern modularizer. We push the network to represent regions from different classes using the same set of region patterns $\{\bm{P}_n\}$, which reduces the bias on seen classes during training. As shown in Fig.~\ref{fig:mod}, we first leverage a region encoder to map $\bm{R}_m$ as a feature $\bm{f}_R$. Then, we concatenate $\bm{f}_R\in\mathbb R^{1\times E}$ to each point of $\bm{P}_n$ to form an intermediate representation with a dimensionality of $P\times(3+E)$ which is further transformed into a modularization $\bm{r}_m^n\in\mathbb{R}^{P\times3}$. Finally, we concatenate all modularization from different patterns into one pattern modularized region $\bm{R}_m'\in\mathbb{R}^{NP\times 3}$. Note that we also conduct this pattern modularization procedure in local region coordinate system, so we translate the pattern modularized regions $\bm{R}_m'$ back to object-centered coordinate system by reversing the centering procedure defined in Eq.~\ref{eq:c2d} below,

\begin{equation}
\label{eq:c2dback}
\begin{aligned}
\bm{R}_m'\gets\bm{R}_m'+\bm{c}_m.
\end{aligned}
\end{equation}

\begin{wrapfigure}[9]{r}{0.45\linewidth}
\includegraphics[width=\linewidth]{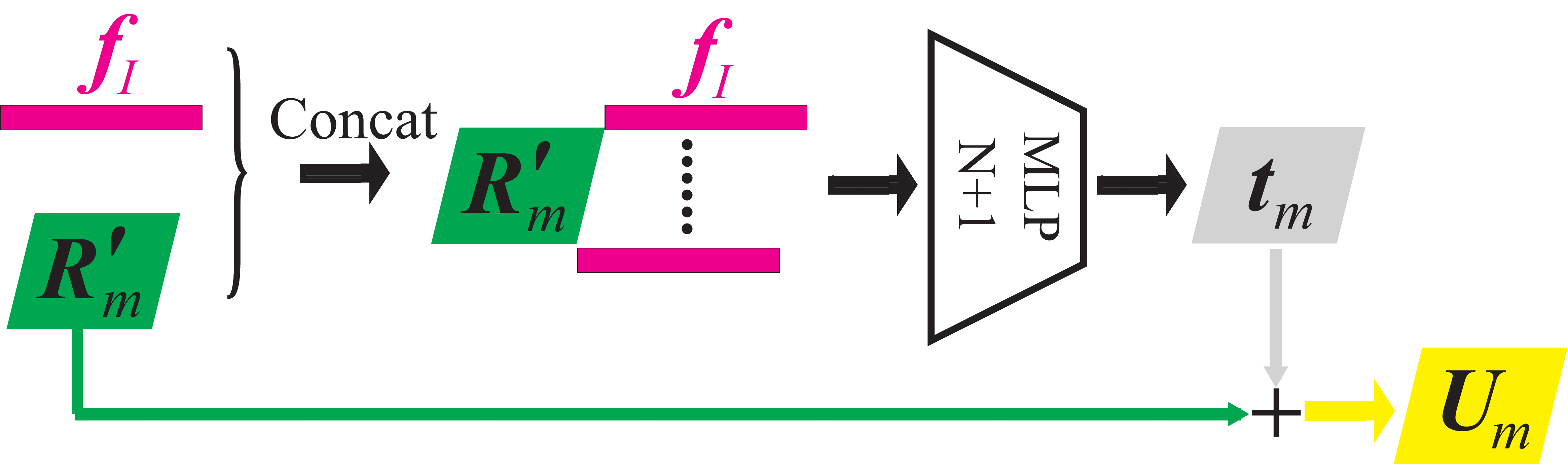}
\caption{\label{fig:cus}The architecture of modularization customizer.}
\end{wrapfigure}

\noindent\textbf{Modularization Customizer. }Based on the pattern modularized region $\bm{R}_m'$, we further customize it using the input image $\bm{I}$ in a modularization customizer. Our purpose is to get more detailed geometry from $\bm{I}$ since $\bm{R}_m'$ merely covers a coarse structure of local regions. Our solution is to push the network to learn how to adjust a region using the content in the image accordingly, which further increases the generalization ability of the local prior. We demonstrate modularization customizer in Fig.~\ref{fig:cus}. We leverage the idea of ResNet~\cite{DBLPHeZRS16} to predict the modularization shift $\bm{t}_m\in\mathbb{R}^{NP\times 3}$ for each $\bm{R}_m'$. We concatenate the latent code $\bm{f}_I$ of the image $\bm{I}$ to each point of $\bm{R}_m'$, which forms an intermediate representation with a dimensionality of $NP\times (3+H)$. This intermediate representation is further transformed into a modularization shift $\bm{t}_m$ by an MLP. Finally, we got the pattern customized region $\bm{U}_m$ below,

\begin{equation}
\label{eq:cc4}
\begin{aligned}
\bm{U}_m=\bm{R}_m'+\bm{t}_m.
\end{aligned}
\end{equation}

We reconstruct the final point cloud $\bm{F}\in\mathbb{R}^{F\times 3}$ by concatenating all pattern customized regions $\{\bm{U}_m, m\in[1,M]\}$ together. Note that we remove the points on each $\bm{U}_m$ that have the same indexes of zero points padded to $\bm{R}_m$ to reduce the redundance. To regulate the pattern customized region $\bm{U}_m$ in a specific region, we add a local shape constraint to minimize the CD distance between each $\bm{U}_m$ and the corresponding GT region $\bm{G}_m$,

\begin{equation}
\label{eq:cd2}
\begin{aligned}
L_{Region}=\frac{1}{M}\sum_{m=1}^{m=M}&\sum_{g\in\bm{G}_m}\min_{u\in\bm{U}_m}||u-g||_2+\\
&\sum_{u\in\bm{U}_m}\min_{g\in\bm{G}_m}||u-g||_2.
\end{aligned}
\end{equation}

\section{Experiments and Anaylysis}
We evaluate our performance by comparing our method with the state-of-the-art ones in point cloud reconstruction from seen classes and unseen classes.
\subsection{Setup}
\noindent\textbf{Details. }To highlight the effectiveness of our idea, we leverage a simple neural network in our experiments. We use a network introduced in Differentiable Point Clouds~\cite{InsafutdinovD18} as 2D encoder and shape decoder in Fig.~\ref{fig:Overview}. Each one of the $N$ pattern learners is an MLP with 3 fully connected layers. In pattern modularizer in Fig.~\ref{fig:mod}, the region encoder is a fully connected layer, and each one of the $N$ MLP has 4 fully connected layers, while the MLP is formed by 3 fully connected layers in modularization customizer in Fig.~\ref{fig:cus}.

We reconstruct initial shape prediction $\bm{S}$ and final reconstruction $\bm{F}$ as $S=F=2048$ points. We split $\bm{S}$ into $M=8$ regions. Each region is represented by $N=8$ patterns, and each pattern is formed by $P=256$ points. The feature $\bm{f}_I$ of the input image $\bm{I}$ is $H=1024$ dimensional, while the feature $\bm{f}_R$ of region $\bm{R}_m$ is $E=64$ dimensional.

\noindent\textbf{Dataset and Metric. }For fair comparison with the state-of-the-art methods, we conduct experiments using ShapeNet~\cite{ChangFGHHLSSSSX15} and Pixel3D~\cite{pix3d} under different experiment conditions. In numerical comparison, we will elaborate on the experiment conditions including the classes used during training and test and the number of points used in evaluation. Moreover, we employ L1-CD defined in Eq.~\ref{eq:cd} and IoU to evaluate the results. The results of L1-CD are produced by comparing our reconstruction and the ground truth with the same number of points. Our results of IoU are produced using voxel grids obtained by the method introduced in DPC~\cite{InsafutdinovD18} at a resolution of $32^3$ which keeps the same as others.

\begin{table}[thb]
\centering
\resizebox{\linewidth}{!}{
    \begin{tabular}{c|c|c|c|c|c|c|c|c|c|c|c|c|c|c||c}  
     \hline
        &Methods& Airplane& Bench & Cabinet & Car &Chair&Display& Lamp &Speaker&Rifle&Sofa&Table&Phone&Vessel&Mean \\  
     \hline
     \multirow{9}{*}{\rotatebox{90}{CD}}&R2N2&0.227&0.194&0.217&0.213&0.270&0.605&0.778&0.318&0.183&0.229&0.239&0.195&0.238&0.278\\
     &PSGN&0.137&0.181&0.215&0.169&0.247&0.284&0.314&0.316&0.134&0.224&0.222&0.161&0.188&0.215\\
     &Pix2Mesh&0.187&0.201&0.196&0.180&0.265&0.239&0.308&0.285&0.164&0.212&0.218&0.149&0.212&0.216\\
     &AtlasNet&0.104&0.138&0.175&0.141&0.209&0.198&0.305&0.245&0.115&0.177&0.190&0.128&0.151&0.175\\
     &OccNet&0.134&0.150&0.153&0.149&0.206&0.258&0.368&0.266&0.143&0.181&0.182&0.127&0.201&0.194\\
     &3D43D&0.096&0.112&0.119&0.122&0.193&0.166&0.561&0.229&0.248&0.125&0.146&0.107&0.175&0.184\\
     &GraphX&0.024&0.037&0.039&0.033&0.047&0.050&0.048&0.054&0.026&0.057&0.051&0.024&0.037&0.041\\
     &SDT&0.042&0.034&0.049&0.029&\textbf{0.036}&0.047&0.062&0.064&0.054&\textbf{0.041}&\textbf{0.033}&0.032&0.038&0.039\\
     \cline{2-16}
     &Ours&\textbf{0.019}&\textbf{0.032}&\textbf{0.037}&\textbf{0.027}&0.040&\textbf{0.046}&\textbf{0.043}&\textbf{0.046}&\textbf{0.018}&0.049&0.044&\textbf{0.020}&\textbf{0.033}&\textbf{0.035}\\
     \hline
     \multirow{5}{*}{\rotatebox{90}{IoU}}&R2N2&0.561&0.527&0.772&0.836&0.550&0.565&0.421&0.717&0.600&0.706&0.580&0.754&0.610&0.631\\
     &PSGN&0.601&0.550&0.771&0.831&0.544&0.552&0.462&0.737&0.604&0.708&0.606&0.749&0.611&0.640\\
     &GAL&0.685&0.709&0.772&0.737&0.700&0.804&0.670&0.698&0.715&0.739&0.714&0.773&0.675&0.712\\
     &GraphX&0.791&0.746&0.770&0.821&0.704&0.765&0.573&0.715&0.765&0.786&0.688&\textbf{0.848}&0.772&0.750\\
     \cline{2-16}
     &Ours&\textbf{0.802}&\textbf{0.765}&\textbf{0.808}&\textbf{0.841}&\textbf{0.715}&\textbf{0.812}&\textbf{0.679}&\textbf{0.746}&\textbf{0.780}&\textbf{0.790}&\textbf{0.732}&0.844&\textbf{0.783}&\textbf{0.776}\\
   \hline
   \end{tabular}}
    \caption{Accuracy of reconstruction with 2048 points under ShapeNet for seen classes in terms of L1-CD and IoU.}  
   \label{table:t10}
\end{table}

\subsection{Reconstruction from Seen Classes}
\noindent\textbf{Numerical Evaluation. }We first evaluate our method under all 13 seen classes in ShapeNet dataset. We train our model using the training set of all 13 classes, while testing the trained model using the test set from the 13 seen classes. We compare our method with the latest methods designed for different 3D representations, including voxel based method R2N2~\cite{ChoyXGCS16}, mesh based methods Pix2Mesh~\cite{WangZLFLJ18}, point cloud based methods PSGN~\cite{FanSG17}, AtlasNet~\cite{Groueix_2018_CVPR}, GraphX~\cite{DBLP:conf/iccv/NguyenCK019}, and SDT~\cite{Hu2019LearningTG}, and implicit function based method OccNet~\cite{MeschederNetworks} and 3D43D~\cite{bautista2020generalization}. We report our numerical comparison under each one of 13 classes in Table~\ref{table:t10}. The comparison demonstrates that our method outperforms other methods in shape reconstruction. Similarly, our IoU comparison with R2N2~\cite{ChoyXGCS16}, PSGN~\cite{FanSG17}, GAL~\cite{DBLP:conf/eccv/JiangSQJ18}, and GraphX~\cite{DBLP:conf/iccv/NguyenCK019} in Table~\ref{table:t10} also shows that our method can reveal more accurate structures in reconstructions.


\begin{wrapfigure}[33]{r}{0.55\linewidth}
\includegraphics[width=\linewidth]{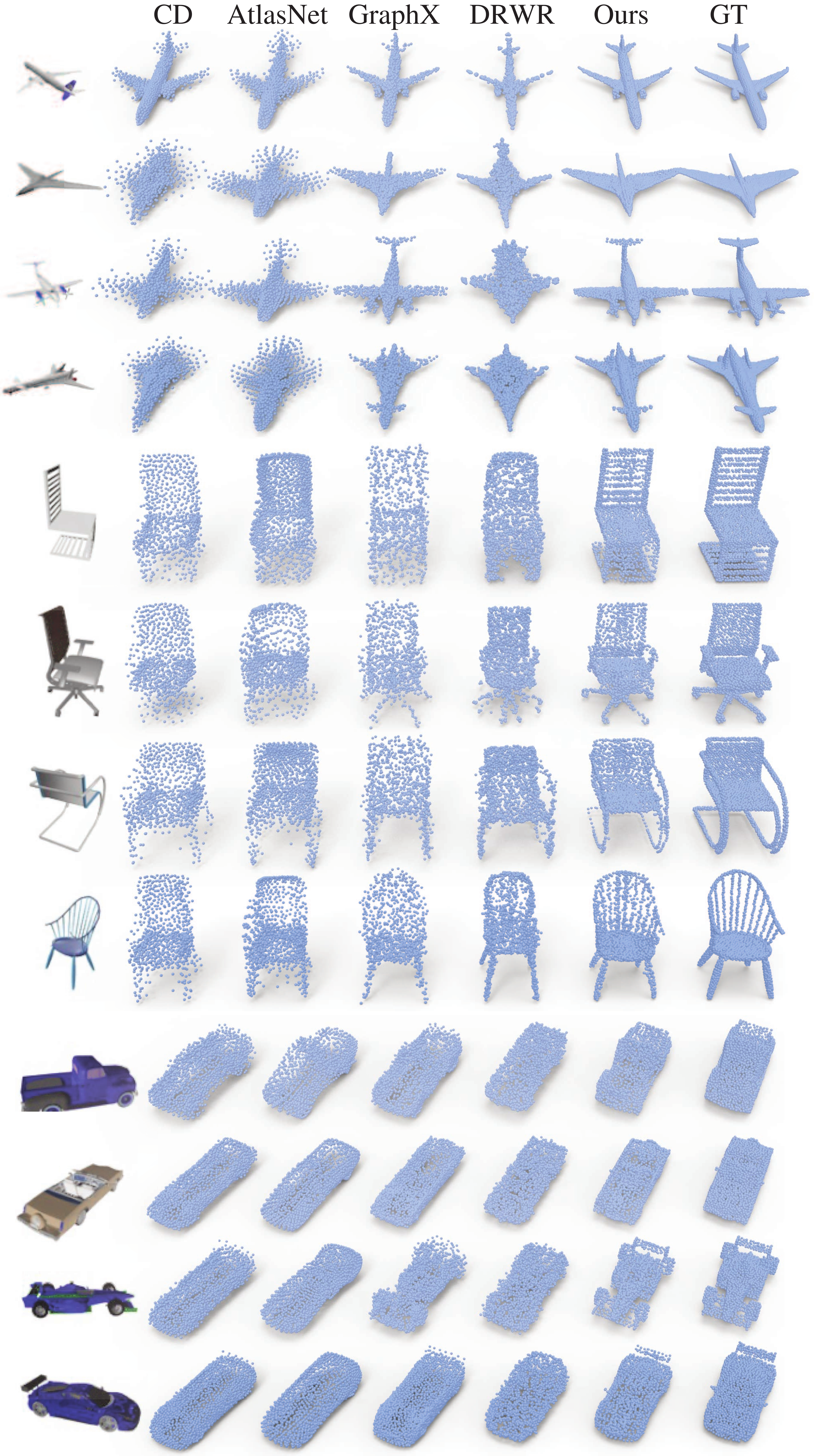}
\caption{\label{fig:seencom}The visual comparison under seen classes in ShapeNet.}
\end{wrapfigure}

Additionally, we compare with some methods that are just trained under a subset of ShapeNet dataset which includes Chair, Car, Plane, Table and Motorcycle. To highlight our advantage, we train our model only using 3 classes including Chair, Car, and Plane, but test our model under all the 5 classes, which keeps the same as others. For fair comparison, we down sample our reconstruction results to 1024 points to compare it with the ground truth, which keeps the same as our compared methods.

We report our average accuracy over the 5 classes by comparing with viewer-centered methods and object-centered methods for different 3D representations in the ``Seen'' column in Table~\ref{table:NOX11}. The viewer-centered methods, including DRC~\cite{TulsianiZEM17}, MarrNet~\cite{DBLP:conf/nips/0001WXSFT17}, GenRe~\cite{NIPS2018_7494}, GSIR~\cite{DBLP:conf/eccv/WangF20}, reconstruct shapes in camera coordinate system, which require camera poses to align ground truth shapes to the images. This makes it hard to train neural network to converge to high accurate reconstructions, but the network will be more generalized to unseen classes~\cite{NIPS2018_7494}. While object-centered methods, including IMNet~\cite{chen2018implicit_decoder}, OccNet~\cite{MeschederNetworks}, DeepSDF~\cite{Park_2019_CVPR}, AtlasNet~\cite{Groueix_2018_CVPR}, DRWR~\cite{handrwr2020}, can reconstruct more accurate shapes in canonical coordinate system. Our method not only achieves the best performance among all object-centered methods even we are using much less seen classes during training, but also generalized better to unseen classes. Note that we reproduce the results of DRWR by training it under the same 3 classes as ours using its code.

\noindent\textbf{Visual Comparison. }We visually compare our method with the state-of-the-art in Fig.~\ref{fig:seencom}. We can see that our method can reveal more accurate geometry than others, where we also show our baseline reconstruction as ``CD'' which is obtained by training 2D encoder and shape decoder merely using the $L_{Shape}$ loss.

\subsection{Reconstruction from Unseen Classes}

\noindent\textbf{Evaluation in ShapeNet. }We first evaluate our trained model which produces the seen results under ShapeNet in Table~\ref{table:NOX11} under 4 unseen classes including Bench, Sofa, Bed, and Vessel. 

\begin{wraptable}[13]{r}{0.4\linewidth}
\centering
\resizebox{\linewidth}{!}{
   \begin{tabular}{c|c|c|c}  
    \hline
    \multicolumn{2}{c|}{Method}&Seen &Unseen\\
    \hline
    \multirow{4}{*}{Viewer-Centered}&DRC&0.0970&0.1270\\
    &MarrNet&0.0810&0.1160\\
    &GenRe&0.0680&0.1080\\
    &GSIR&0.0680&0.0990\\
   \hline
   \hline
   \multirow{6}{*}{Object-Centered}&IMNet&0.0550&0.1190\\
    &OccNet&0.0600&0.1280\\
    &DeepSDF&0.0530&0.1150\\
    &AtlasNet&0.0630&0.1260\\
    &DRWR&0.0536&0.0715\\
    \cline{2-4}
    &Ours&\textbf{0.0527}&\textbf{0.0540}\\
    \hline
  \end{tabular}}
  \caption{L1-CD accuracy of reconstruction with $1024$ points.}  
  \label{table:NOX11}
\end{wraptable}
We report our average reconstruction accuracy with 1024 points compared to the ground truth point clouds in the ``Unseen'' column in Table~\ref{table:NOX11}, where we down sample our reconstruction with 2048 points to 1024 points. The comparison shows that our method can significantly outperform the other methods which learn a global prior for shape reconstruction from unseen classes. Moreover, our method also shows much better generalization ability than GSIR~\cite{DBLP:conf/eccv/WangF20} which aims to generalize a learned global prior.
The superior over GSIR~\cite{DBLP:conf/eccv/WangF20} justifies that our idea of generalizing local prior of reconstruction is more promising.

\begin{table}[htb]
\centering
\resizebox{\linewidth}{!}{
    \begin{tabular}{c|c|c|c|c|c|c|c|c|c|c|c|c}  
     \hline
     \multicolumn{2}{c|}{Method}&Bench&Vessel&Rifle&Sofa&Table&Phone&Cabinet&Speaker&Lamp&Display&Mean\\
     \hline
     \multirow{4}{*}{Viewer-Centered}&DRC&0.120&0.109&0.121&0.107&0.129&0.132&0.142&0.141&0.131&0.156&0.129\\
     &MarrNet&0.107&0.094&0.125&0.090&0.122&0.117&0.125&0.123&0.144&0.149&0.120\\
     &Multi-View&0.092&0.092&0.102&0.085&0.105&0.110&0.119&0.117&0.142&0.142&0.111\\
     &GenRe&0.089&0.092&0.112&0.082&0.096&0.107&0.116&0.115&0.124&0.130&0.106\\
    \hline
    \hline
    \multirow{6}{*}{Object-Centered}&DRC&0.112&0.100&0.104&0.108&0.133&0.199&0.168&0.164&0.145&0.188&0.142\\
     &AtlasNet&0.102&0.092&0.088&0.098&0.130&0.146&0.149&0.158&0.131&0.173&0.127\\
     &GraphX&0.111&0.065&0.119&0.098&0.138&0.120&0.113&0.111&0.134&0.114&0.112\\
     &DRWR&0.075&0.059&0.104&0.070&0.100&0.094&0.088&0.086&0.102&0.097&0.088\\
     &CD&0.110&0.084&0.121&0.122&0.114&0.136&0.126&0.122&0.143&0.160&0.124\\
     \cline{2-13}
     &Ours&\textbf{0.054}&\textbf{0.046}&\textbf{0.046}&\textbf{0.058}&\textbf{0.070}&\textbf{0.061}&\textbf{0.071}&\textbf{0.072}&\textbf{0.089}&\textbf{0.077}&\textbf{0.064}\\
     \hline
   \end{tabular}}
   \caption{L1-CD accuracy of reconstruction with $1024$ points for unseen classes.}  
   \label{table:NOX21}
\end{table}

\begin{table*}[htb]
\centering
\resizebox{0.85\linewidth}{!}{
    \begin{tabular}{c|c|c|c|c|c|c|c|c|c|c||c}  
     \hline
     \multicolumn{1}{c|}{Method}&Bench&Vessel&Rifle&Sofa&Table&Phone&Cabinet&Speaker&Lamp&Display&Mean\\
     \hline
     3D43D&0.357&0.521&0.707&0.421&0.583&0.996&0.529&0.744&1.997&1.389&0.824\\
     SDFNet&0.133&0.209&0.199&0.306&0.288&0.434&0.241&0.374&0.554&0.487&0.323\\
     HPN&0.079&0.071&0.070&0.144&0.148&0.064&0.114&0.110&0.147&0.163&0.111\\
     Point-e&0.084&0.155&0.103&0.100&0.135&0.207&0.102&0.104&0.195&0.112&0.130\\
     \hline
     Ours&\textbf{0.049}&\textbf{0.042}&\textbf{0.042}&\textbf{0.051}&\textbf{0.064}&\textbf{0.054}&\textbf{0.062}&\textbf{0.063}&\textbf{0.082}&\textbf{0.070}&\textbf{0.058}\\
     \hline
   \end{tabular}}
   \caption{L1-CD accuracy of reconstruction with $2048$ points for unseen classes.}  
   \label{table:NOX31}
\end{table*}

Then, we report our numerical comparison under more unseen classes in ShapeNet. In this experiment, we also use our model trained under Chair, Plane, Car, in Table~\ref{table:NOX11}, while testing under 10 unseen classes shown in Table~\ref{table:NOX21}. We conduct the comparison with viewer-centered methods including MarrNet~\cite{DBLP:conf/nips/0001WXSFT17}, Multi-View~\cite{DBLP:conf/cvpr/ShinFH18}, GenRe~\cite{NIPS2018_7494}, and object-centered methods including DRC~\cite{TulsianiZEM17}, AtlasNet~\cite{Groueix_2018_CVPR}, GraphX~\cite{DBLP:conf/iccv/NguyenCK019}, DRWR~\cite{handrwr2020}. 


\begin{wrapfigure}[22]{r}{0.5\linewidth}
\includegraphics[width=\linewidth]{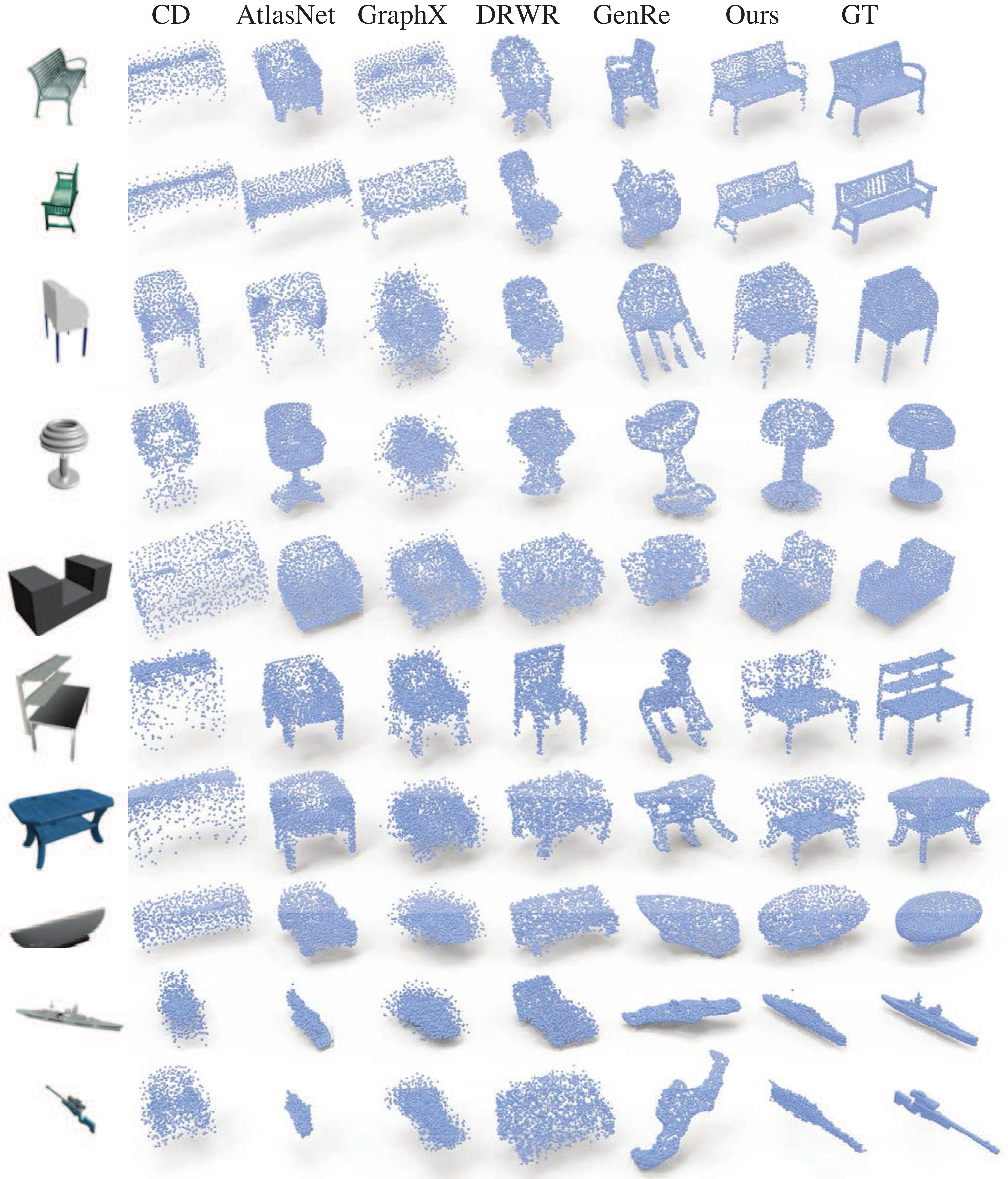}
\caption{\label{fig:unseencom}The comparison under unseen classes in ShapeNet.}
\end{wrapfigure}

The comparison shows that our learned local prior for shape reconstruction can generalize to unseen classes better than the methods learning global prior including AtlasNet, GraphX, DRWR. Although our model is trained using ground truth point clouds in object-centered coordinate system, it still generalizes to unseen classes with higher reconstruction accuracy than viewer-centered methods including MarrNet, Multi-View, and GenRe. In addition, we also highlight our learned local prior by comparing our final reconstruction $\bm{F}$ with the initial shape prediction $\bm{S}$ obtained by merely using $L_{Shape}$ in Eq.~(\ref{eq:cd}). Our significant improvement over the results of ``CD'' demonstrates that learning local prior is very helpful for the reconstruction generalization to unseen classes.


\begin{wrapfigure}[12]{r}{0.35\linewidth}
\includegraphics[width=\linewidth]{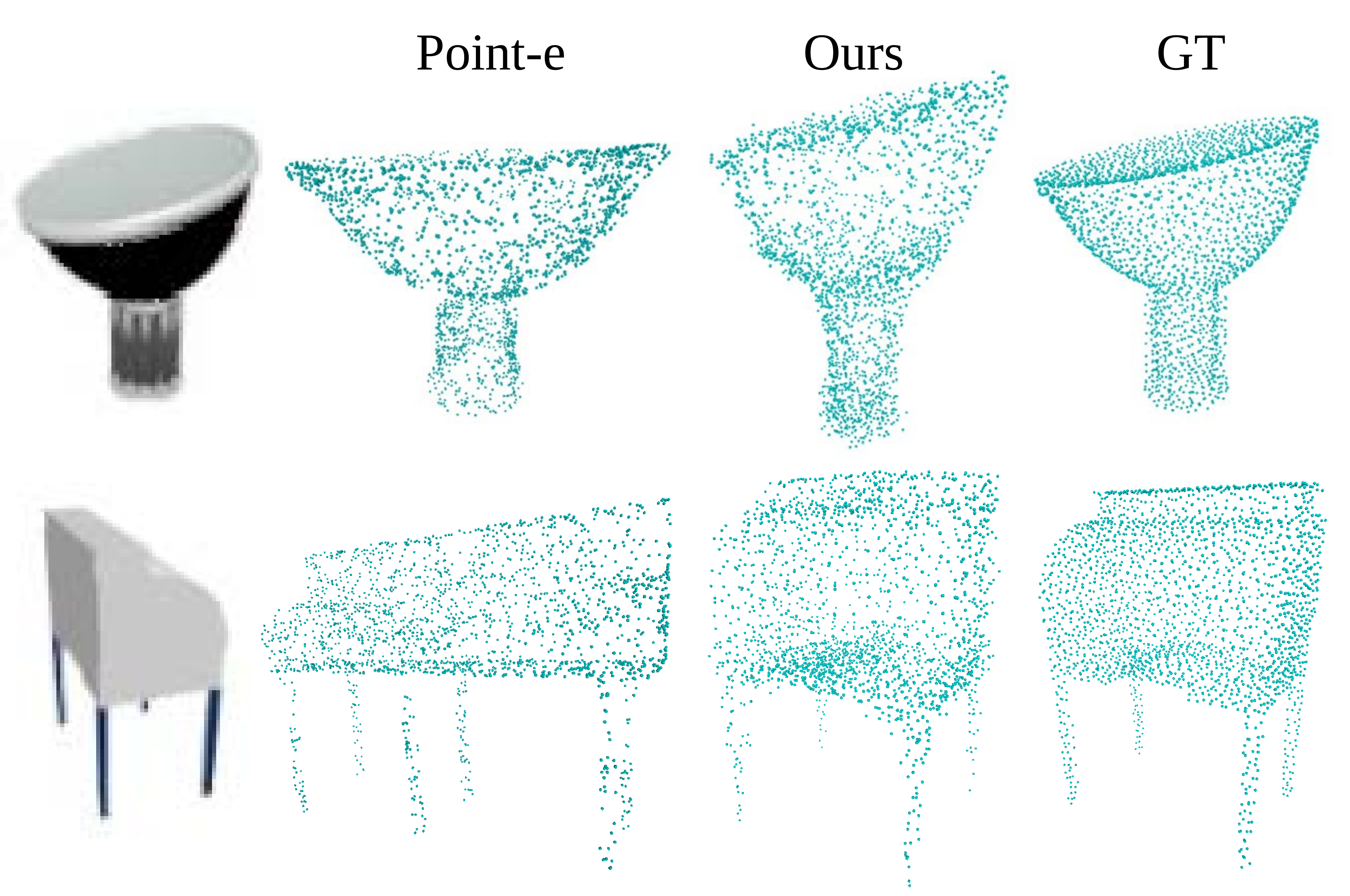}
\caption{\label{fig:unseencom1}The comparison under unseen classes with Point-e.}
\end{wrapfigure}

Under the same condition, we report the comparison with methods learning implicit functions, including the viewer-centered method 3D43D~\cite{bautista2020generalization}, SDFNet~\cite{3DVUnseen}, and HPN~\cite{Bechtold2021HPN}, and Point-E~\cite{nichol2022point} for $2048$ point reconstruction.
The comparison in Table~\ref{table:NOX31} shows that our method significantly outperforms these methods, although they require lots of queries sampled in 3D space to learn implicit functions, which is more detailed supervision than our surface points.


\noindent\textbf{Evaluation in Pixel3D. }Finally, we evaluate our generalization performance in Pixel3D dataset. We leverage our model trained under Chair, Plane, and Car in ShapeNet in Table~\ref{table:NOX11} to reconstruct shapes from 5 unseen classes in Pixel3D dataset, including Bed, Bookcase, Desk, Sofa, and Wardrobe. We compare object-centered methods for point clouds reconstruction including Atlasnet~\cite{Groueix_2018_CVPR}, GraphX~\cite{DBLP:conf/iccv/NguyenCK019}, viewer-centered methods from unseen classes including GenRe~\cite{NIPS2018_7494}, GSIR~\cite{DBLP:conf/eccv/WangF20}, and also our initial shape prediction (``CD'') with merely $L_{Shape}$ as loss. The comparison in Table~\ref{table:NOX41} also demonstrates our superior over the state-of-the-art.


\begin{wrapfigure}[24]{r}{0.5\linewidth}
\includegraphics[width=\linewidth]{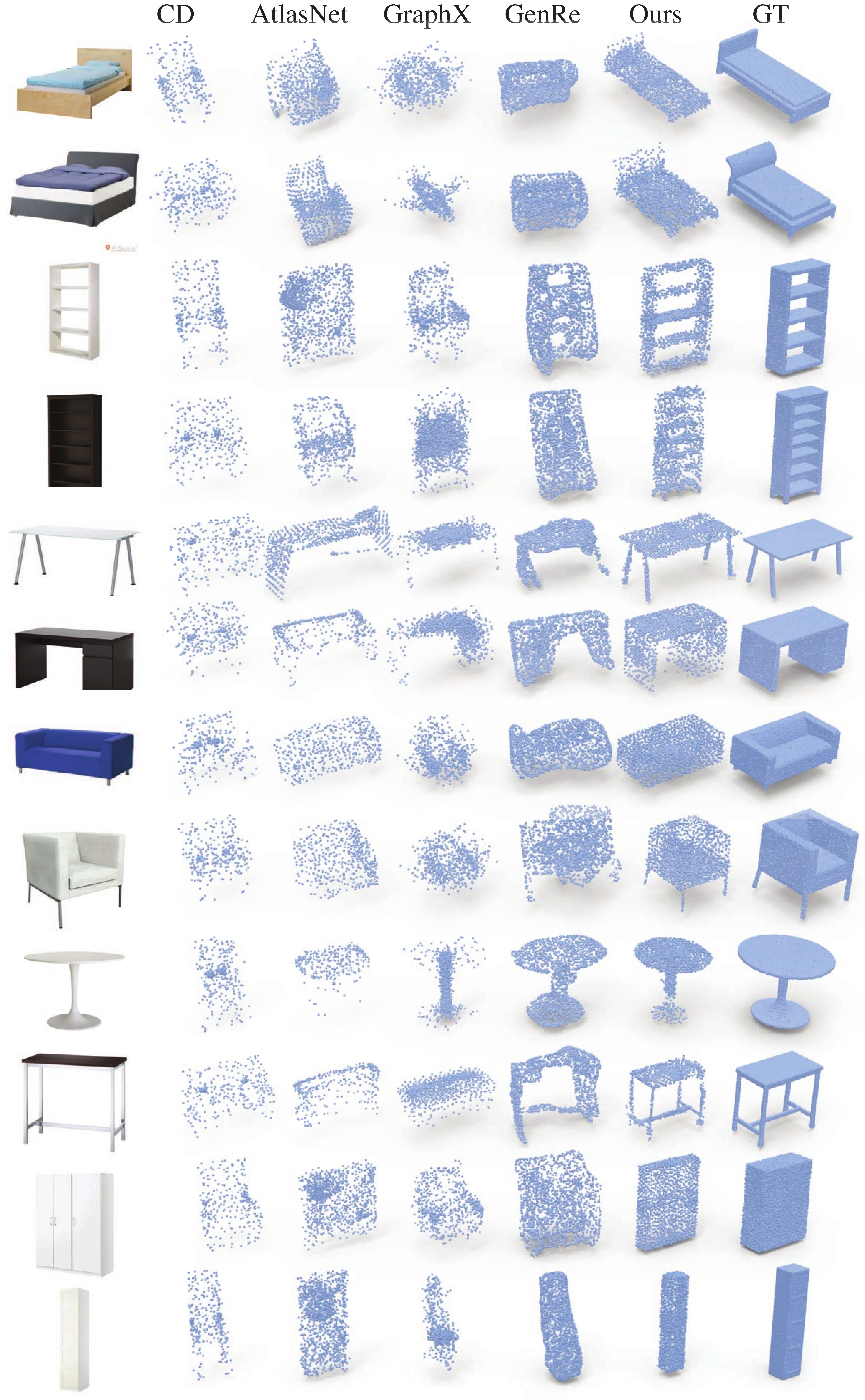}
\caption{\label{fig:unseencom1}The visual comparison under unseen classes in Pixel3D.}
\end{wrapfigure}

\begin{figure}[tb]
  \centering
   \includegraphics[width=0.8\linewidth]{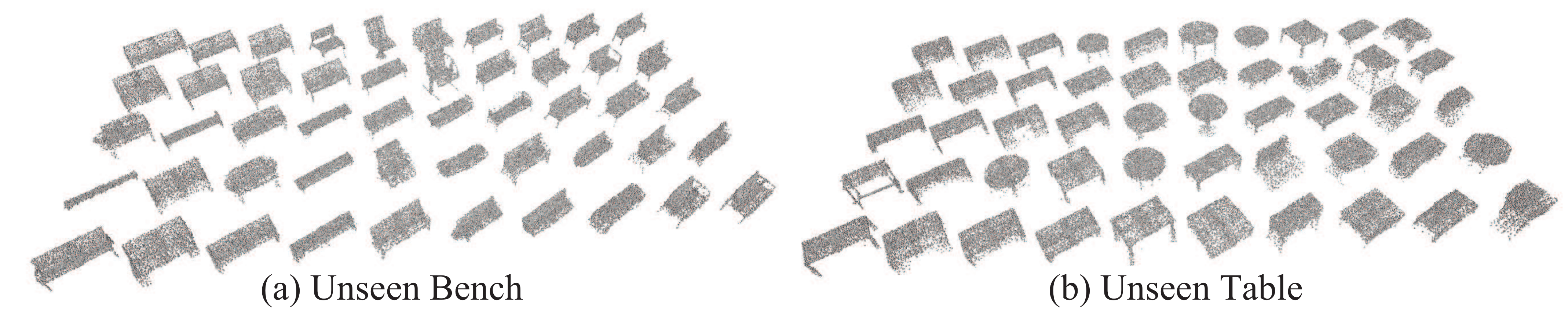}
  %
  %
\caption{\label{fig:unseencommore}More reconstruction from unseen classes under (a) ShapeNet and (b) Pixel3D.}
\end{figure}

\noindent\textbf{Visual Comparison. }Our visual comparison with the state-of-the-art under unseen classes in ShapeNet is shown in Fig.~\ref{fig:unseencom} and Fig.~\ref{fig:unseencom1}.
It demonstrates that the compared methods do not generalize well to unseen classes to reconstruct plausible shapes, such as the baseline CD, AtlasNet, GraphX, and DRWR, while GenRe that learns a global prior generalizes to unseen classes with low accuracy in viewer-centered coordinate system. Fig.~\ref{fig:unseencom1} shows our superiority over Point-E~\cite{nichol2022point} which does not perform well on images with occlusion. Our method can leverage the learned local prior to reconstruct more plausible shapes in higher accuracy in object-centered coordinate system. We also conduct a visual comparison under Pixel3D in Fig.~\ref{fig:unseencom1}, which also demonstrates our significant improvements over others. Moreover, we also show more shape reconstructions from unseen classes in Fig.~\ref{fig:unseencommore} under ShapeNet and Pixel3D.


\begin{wraptable}[10]{r}{0.4\linewidth}
\centering
\resizebox{\linewidth}{!}{
   \begin{tabular}{c|c|c|c|c|c}  
    \hline
    Method&Bed&Bookcase&Desk&Sofa&Wardrobe\\
    \hline
    GraphX&0.141&0.122&0.132&0.094&0.116\\
    AtlasNet&0.115&0.137&0.124&0.096&0.119\\
    GenRe&0.111&0.101&0.107&0.085&0.111\\
    GSIR&0.107&0.095&0.100&0.083&0.103\\
    CD&0.165&0.102&0.163&0.104&0.132\\
    \hline
    Ours&\textbf{0.085}&\textbf{0.094}&\textbf{0.089}&\textbf{0.074}&\textbf{0.067}\\
    \hline
  \end{tabular}}
  \caption{L1-CD accuracy of reconstruction with $2048$ points for unseen classes under Pixel3D.}  
  \label{table:NOX41}
\end{wraptable}

\subsection{Analysis}
We conduct experiments under Bench class in ShapeNet, we reconstruct point clouds from seen classes with $2048$ points.

\noindent\textbf{Ablation Studies. }We conduct ablation studies to justify the effectiveness of the elements in our model. We first highlight our local prior. We only use the 2D encoder and point decoder to minimize $L_{Shape}$ in training, and report the result as ``No local'' in Table~\ref{table:NOX51}. The degenerated results indicate that the local prior is important to improve the reconstruction accuracy. Similarly, we explore the effectiveness of pattern modularization by removing the pattern learner and pattern modularization, and the effectiveness of modularization shift by removing the customization procedure, respectively. 

\begin{wraptable}[6]{r}{0.55\linewidth}
\centering
\resizebox{\linewidth}{!}{
   \begin{tabular}{c|c|c|c|c|c}  
    \hline
    No local&No patterns&No shift&No $L_{Region}$&No $L_{Shape}$&Ours\\
    \hline
    0.054&0.049&0.041&0.038&0.048&\textbf{0.032}\\
    \hline
  \end{tabular}}
  \caption{Ablation studies in terms of L1-CD.}  
  \label{table:NOX51}
\end{wraptable}
The results of ``No patterns'' and ``No shift'' degenerates, which indicates that the network requires region patterns and their adjustment to reconstruct various regions on different shapes. We also evaluate the effect of $L_{Region}$ by replacing it into another $L_{Shape}$. The result of ``No $L_{Region}$'' justifies that $L_{Region}$ is important for the detailed local structures. Although we aim to learn a local prior, the weak global prior captured by initial shape reconstruction $\bm{S}$ is also helpful to provide the network a good start, as shown by the degenerated results of ``No $L_{Shape}$''.

\begin{wraptable}[6]{r}{0.45\linewidth}
\centering
\resizebox{\linewidth}{!}{
   \begin{tabular}{c|c|c}  
    \hline
    &Sampling on 2D plane&Sampling in 3D voxel\\
    \hline
    L1CD&0.034&\textbf{0.032}\\
    \hline
  \end{tabular}}
  \caption{Sampling effect.}  
  \label{table:NOX61}
\end{wraptable}

\noindent\textbf{Sampled Points. }We learn region patterns by transforming points sampled from voxel grids, since the sampled points occupy all the space where we hold local regions in local coordinate system. We compare these sampled points with the ones sampled on a 2D plane which is introduced to reconstruct 3D patches~\cite{DBLP:conf/nips/DeprelleGFKRA19,Groueix_2018_CVPR}. Comparison in Table~\ref{table:NOX61} shows that sampling on 2D plane is harder to be transformed to represent 3D local structures in voxel grids.


\begin{wraptable}[5]{r}{0.2\linewidth}
\centering
\resizebox{\linewidth}{!}{
   \begin{tabular}{c|c|c|c}  
    \hline
    $M$&1&8&27\\
    \hline
    L1CD&0.035&\textbf{0.032}&0.049\\
    \hline
  \end{tabular}}
  \caption{Region number effect.}  
  \label{table:NOX81}
\end{wraptable}

\noindent\textbf{Region Number $M$. }We explore the effect of region number $M$ by trying different region number candidates including $\{1,8,27\}$.
The comparison in Table~\ref{table:NOX81} demonstrates that it is hard to capture structures in local regions if the regions are too large (``1'') or too small (``27''), both of which results in reconstructions with low accuracy.


\begin{wraptable}[7]{r}{0.43\linewidth}
\centering
\resizebox{\linewidth}{!}{
   \begin{tabular}{c|c|c|c}  
     \hline
     &1 pattern,1 region&8 patterns,8 regions&GraphX\\
     \hline
    Bench &0.038&\textbf{0.032}&0.037\\
    Plane &0.121&\textbf{0.114}&0.121\\
    Car&0.100&\textbf{0.090}&0.095\\
     \hline
   \end{tabular}}
  \caption{Comparison with one region in terms of L1-CD.}  
  \label{table:91}
\end{wraptable}

\noindent\textbf{Pattern Number $N$. }We also report the effect of pattern number $N$ by reconstructing point clouds using $\{2,4,8,16\}$ patterns. The comparison in Table~\ref{table:NOX91} shows that it is adequate to use $N=8$ patterns to represent local regions with $M=8$. Since we have modularization customizer to further adjust the pattern modularized regions, our model also does not require a large number of region patterns. 

\begin{wrapfigure}[8]{r}{0.45\linewidth}
\includegraphics[width=\linewidth]{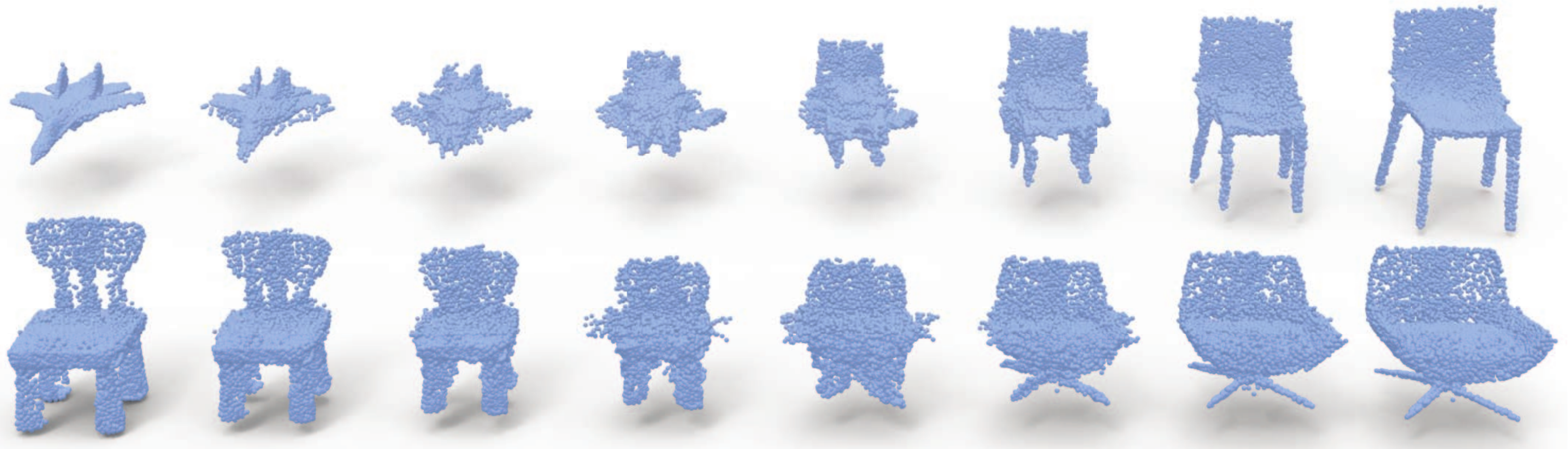}
\caption{\label{fig:interp}The interpolated shapes.}
\end{wrapfigure}

Moreover, we also conduct an experiment to evaluate our generalization ability with only one region (whole shape) and one pattern under seen Bench (training) and unseen Plane and Car (testing). Our method shows much better performance for unseen class reconstruction even with one pattern in Table~\ref{table:91}.


\begin{wraptable}[6]{r}{0.3\linewidth}
\centering
\resizebox{\linewidth}{!}{
   \begin{tabular}{c|c|c|c|c}  
    \hline
    $N$&2&4&8&16\\
    \hline
    L1CD&0.036&0.034&\textbf{0.032}&0.035\\
    \hline
  \end{tabular}}
  \caption{Pattern number effect.}  
  \label{table:NOX91}
\end{wraptable}
\noindent\textbf{Learned Latent Space. }We visualize the learned latent space by reconstructing interpolated shapes from uniformly interpolated latent codes between two point clouds. We use the feature $\bm{f}_I$ of input image to represent each point cloud. The plausible interpolated shapes in Fig.~\ref{fig:interp} demonstrate the semantic meaning of the learned latent space.


\begin{wrapfigure}[10]{r}{0.45\linewidth}
  \centering
   \includegraphics[width=\linewidth]{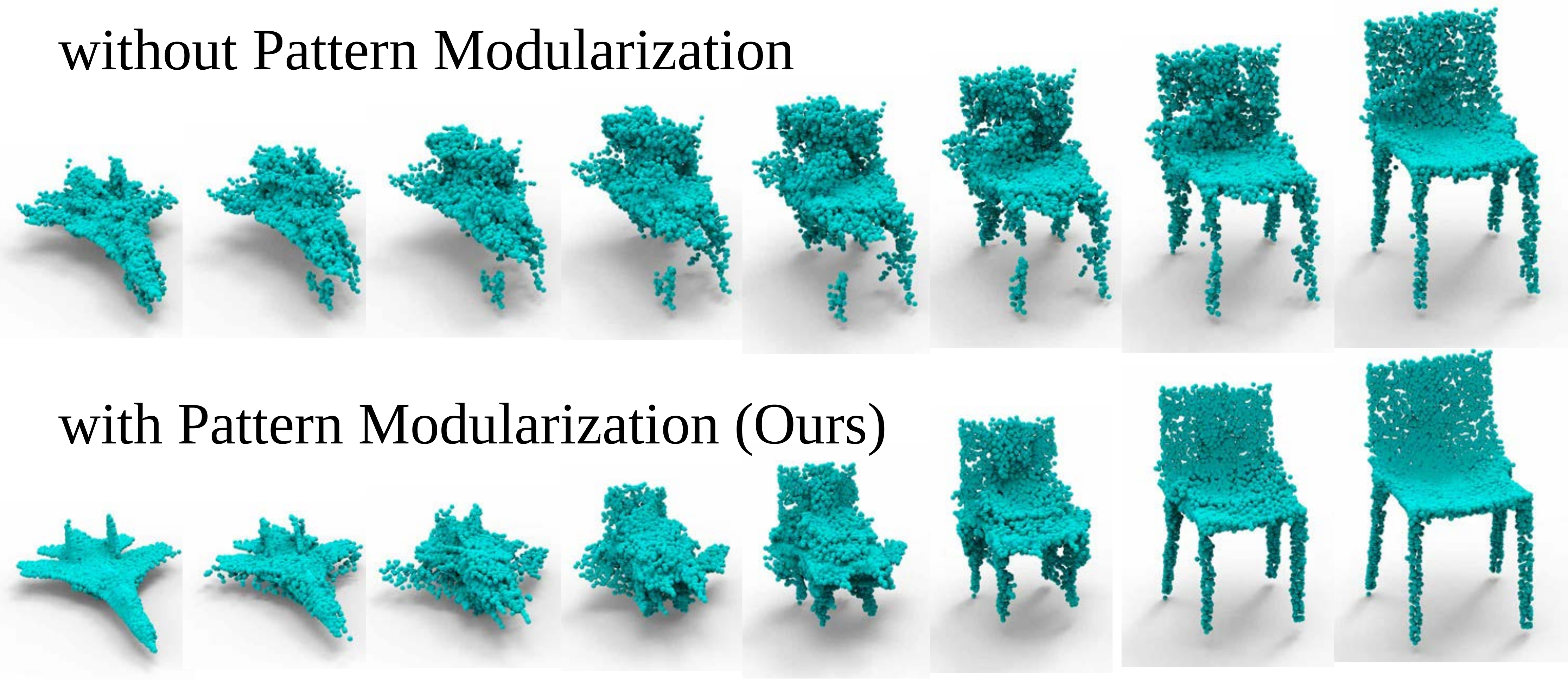}
\caption{\label{fig:1_small}Pattern modularization effect.
}
\end{wrapfigure}

\noindent\textbf{Pattern Modularization. }We visualize the interpolation with or without pattern modularization. As shown in Fig.~\ref{fig:1_small}, It will be hard to learn semantic and meaningful space Without pattern modularization. . We can see that the poorly interpolated shapes without pattern modularization do not show a smooth transition.

\noindent\textbf{Visualization. }We visualize initial shape prediction $\bm{S}$, final reconstruction $\bm{F}$, region patterns $\bm{P}_n$, and pattern modularized regions $\bm{R}_m'$ in Fig.~\ref{fig:pattern}. We can see that $\bm{S}$ produces a coarse shape of the reconstruction,
based on which we reconstruct a more accurate $\bm{F}$ using the learned local prior. All region patterns $\bm{P}_n$ are involved in modularizing each region $\bm{R}_m'$, as color shown, where each pattern represents some structures in the local region and further gets customized to better fit the geometry of a region on $\bm{F}$.

%

\section{Conclusion}
We introduce to reconstruct point clouds from unseen classes by learning local pattern modularization. Our local prior captured by learning and customizing local pattern modularization in seen classes can be effectively generalized to unseen classes in object-centered coordinate system, which leads to much higher reconstruction accuracy. Moreover, our method significantly improves the interpretability of reconstruction from unseen classes using our learned region patterns. We justify the idea of reconstructing regions using only few patterns without requiring any additional information. Our experimental results achieve the state-of-the-art under the widely used benchmarks.

%
%
\bibliographystyle{splncs04}
\bibliography{main}
\end{document}


\title{Supplemental Material for ``Learning Local Pattern Modularization for Point Cloud Reconstruction from Unseen Classes''} 

\titlerunning{Unseen Point Cloud Reconstruction with Local Pattern}




\author{Chao Chen\inst{1} \and
Yu-Shen Liu\inst{1}\thanks{The corresponding author is Yu-Shen Liu. This work was supported by National Key R$\&$D Program of China (2022YFC3800600), the National Natural Science Foundation of China (62272263, 62072268), and in part by Tsinghua-Kuaishou Institute of Future Media Data.} \and
Zhizhong Han\inst{2}}
\authorrunning{C. Chen et al.}
%
\institute{School of Software, BNRist, Tsinghua University, Beijing, China \and
Department of Computer Science, Wayne State University, Detroit, USA\\
\email{chenchao19@mails.tsinghua.edu.cn, liuyushen@tsinghua.edu.cn, h312h@wayne.edu}}

\maketitle

\section{Network and Training}
We leverage the same 2D encoder and shape decoder as DPC~\cite{InsafutdinovD18}. The 2D encoder is a CNN with 7 layers, followed by two fully connected layers of size 1024, and 1024. The shape decoder has 1 fully connected layer of size 3N with tanh to predict a point cloud of N points. The pattern learner consists of three fully connected layers of size 64, 256, and 3 with ReLU on the first two layers and tanh on the final output layer to transform 3D grid points into 3D patterns. The region encoder in pattern modularizer has 1 fully connected layer of size 24 with ReLU, as well as MLP 1 to MLP N consisting of 4 fully connected layers of size 512, 256, 128, 3 with ReLU on the first three layers and tanh on the final output layer. These MLPs do not share parameters with each other. Finally, MLP N+1 in pattern customizer has 3 fully connected layers of size 512, 128, and 3 with ReLU on the first two layers and tanh on the final output layer.

We reconstruct initial shape prediction $\bm{S}$ and final reconstruction $\bm{F}$ as $S=F=2048$ points. We split $\bm{S}$ into $M=8$ regions. Each region is represented by $N=8$ patterns, and each pattern is formed by $P=256$ points. The feature $\bm{f}_I$ of the input image $\bm{I}$ is $H=1024$ dimensional, while the feature $\bm{f}_R$ of region $\bm{R}_m$ is $E=64$ dimensional.

We train our network on an NVIDIA GTX 1080Ti GPU using ADAM optimizer with batch size 4 and an initial learning rate of 0.0001 which is decreased by 0.95 for every 70 epochs. Specifically, our method does not have much computational effort required. a GPU with 5GB memory is enough to run.

\begin{table}[h]
\vspace{-0.0in}
\centering
    \begin{tabular}{c|c|c|c|c}
     \hline
     $\alpha$&0.01&0.1&1&10\\
     \hline
     $L1CD\times100$&5.6075&\textbf{5.5909}&5.9254&5.9531\\
     \hline
   \end{tabular}
   \caption{Effect of $\alpha$.}
   \label{table:weight}
   \vspace{-0.2in}
\end{table}

\section{More Analysis}

\begin{figure}[h]
\centering
    \includegraphics[width=0.9\linewidth]{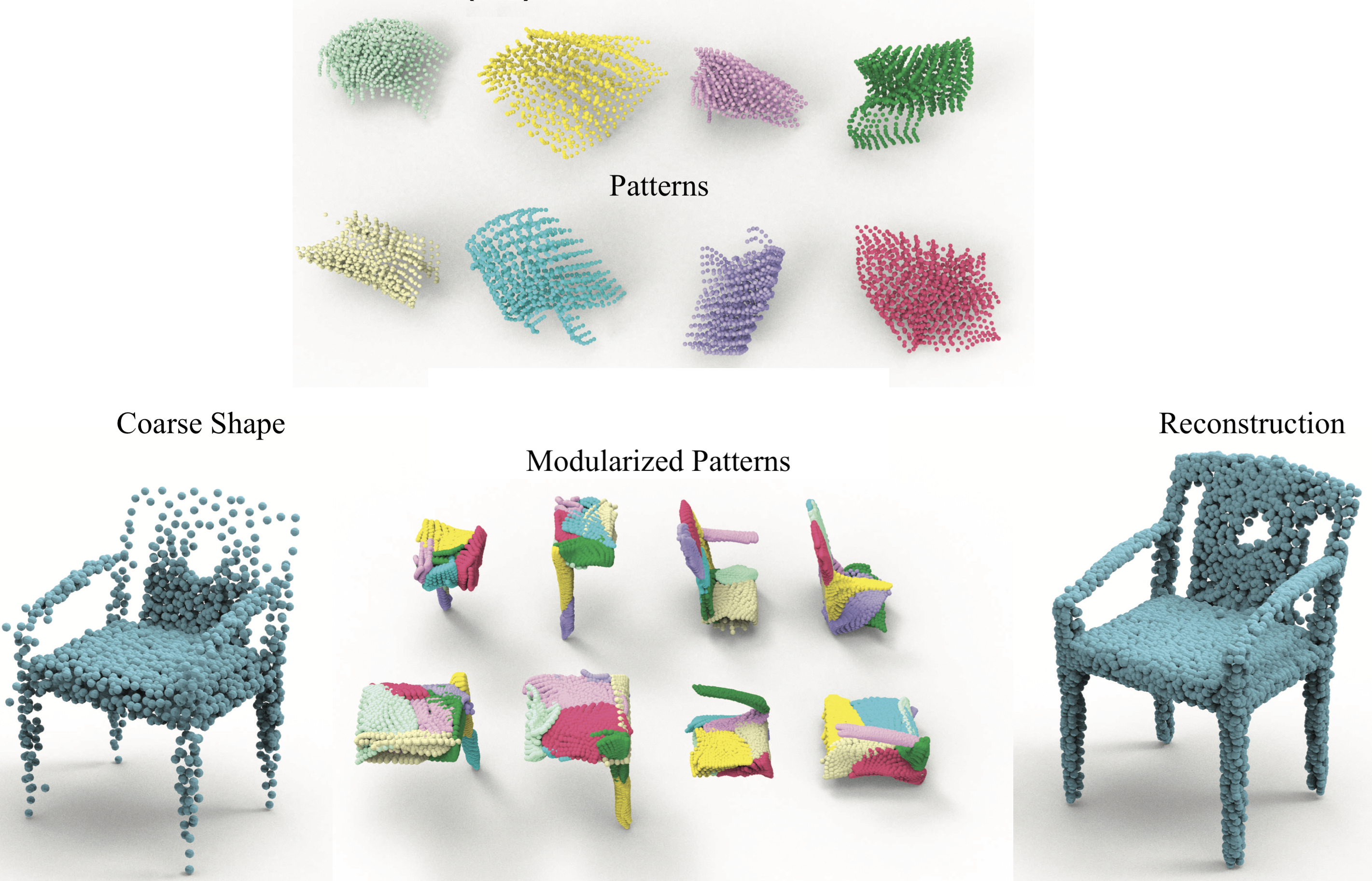}
  \caption{\label{fig:vis}Interpretable visualization.}
\end{figure}

\subsection{Interpretable Visualization}
One key to our local patterns is interpretability since we simplify how the network learns priors, i.e., transforming patterns to form local regions. Fig.1 in our paper visualizes all these transformations with $\{4,8,10,16\}$ patterns, using unique colors to identify a pattern and its transformation. Taking $8$ patterns in Fig.~\ref{fig:vis} for example, we can see how each one of $8$ regions in the middle is formed by the $8$ modularized patterns which are in the same color as the original patterns. We argue that our contributions come from the way how to learn patterns for each region but rather splitting shapes into parts.

\subsection{Loss Weight $\alpha$}
We explore the effect of $\alpha$ balancing the global and local prior learning in Tab.~\ref{table:weight}. The comparisons show that $\alpha=0.1$ is the best. We usually use a small weight on global prior, since local prior can reconstruct accurate shapes in both seen and unseen classes. If $\alpha$ is too large, the network would not learn local prior well, which is not helpful for unseen reconstruction.

\begin{table}[h]
\centering
    \begin{tabular}{c|c|c|c|c|c}  
     \hline
     No local&No patterns&No shift&No $L_{Region}$&No $L_{Shape}$&Ours\\
     \hline
     0.072&0.067&0.061&0.065&0.063&\textbf{0.058}\\
     \hline
   \end{tabular}
   \caption{Ablation studies on unseen classes. }  
   \label{table:unseen}
\end{table}

\subsection{Ablation Studies on Unseen Classes}
We additionally conduct ablation studies on unseen classes to justify the effectiveness of the elements in our model. Similar to Tab.6 in our paper, the related degenerate results in~\ref{table:unseen} including ``No local'', ``No patterns'' and ``No shift'', etc, show that each element in our model helps to generalize the prior well on unseen classes.

\begin{figure}[h]
  \centering
   \includegraphics[width=0.5\linewidth]{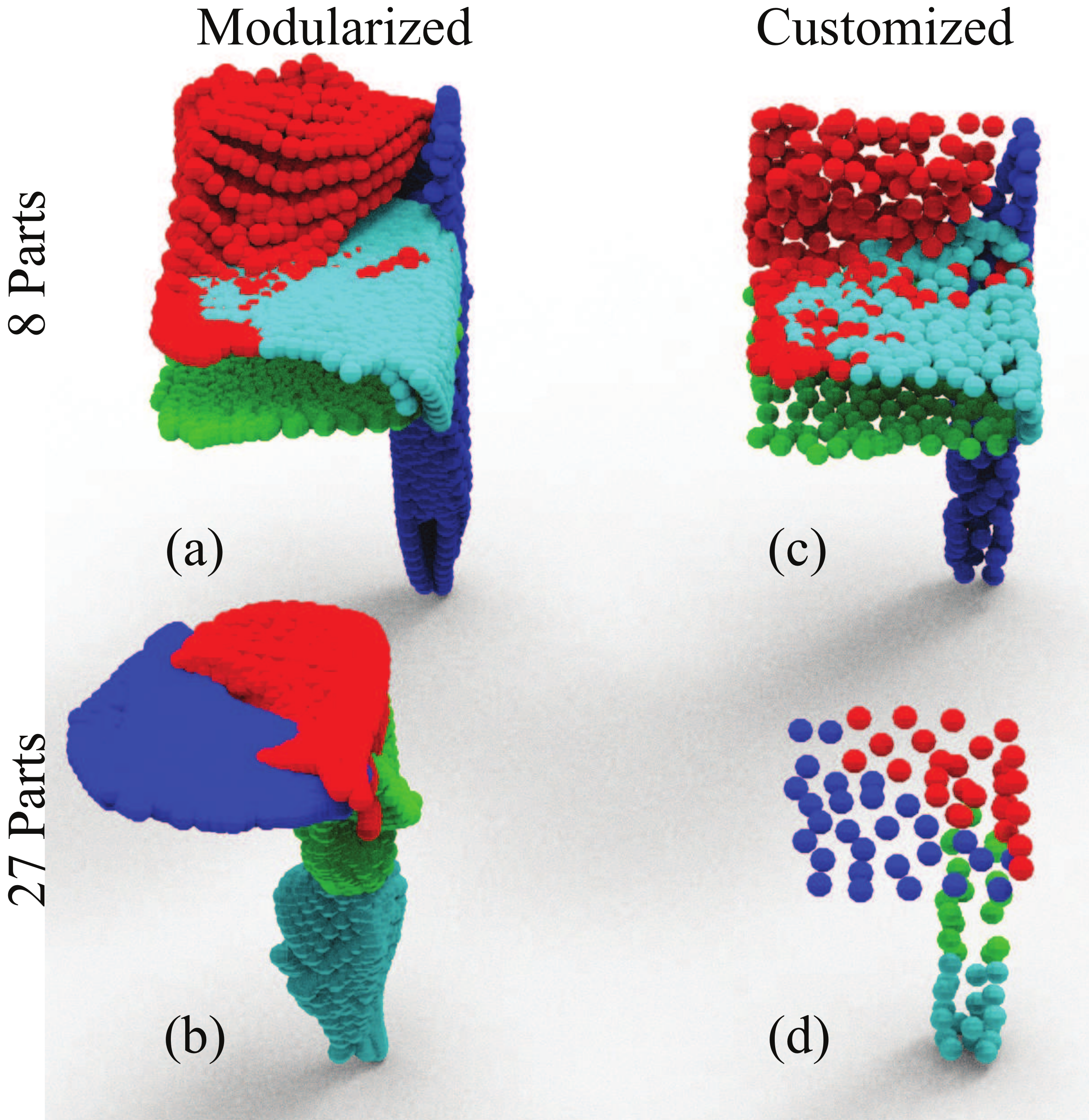}
  %
  %
  \vspace{-0.0in}
\caption{\label{fig:2}Parts Comparison.}
\vspace{-0.2in}
\end{figure}

\subsection{Region Number $M$ }
We demonstrate in our paper that it is hard to capture structures in local regions if the regions are too large (``1'') or too small (``27''), both of which result in reconstructions with low accuracy. The reason why more regions degenerate performance is that more regions result in smaller and sparser regions, which makes it hard for the modularizer to modularize parts with patterns as shown in Fig.~\ref{fig:2}. This further affects learning a good 2D-3D mapping.

\begin{figure}[h]
  \centering
   \includegraphics[width=0.9\linewidth]{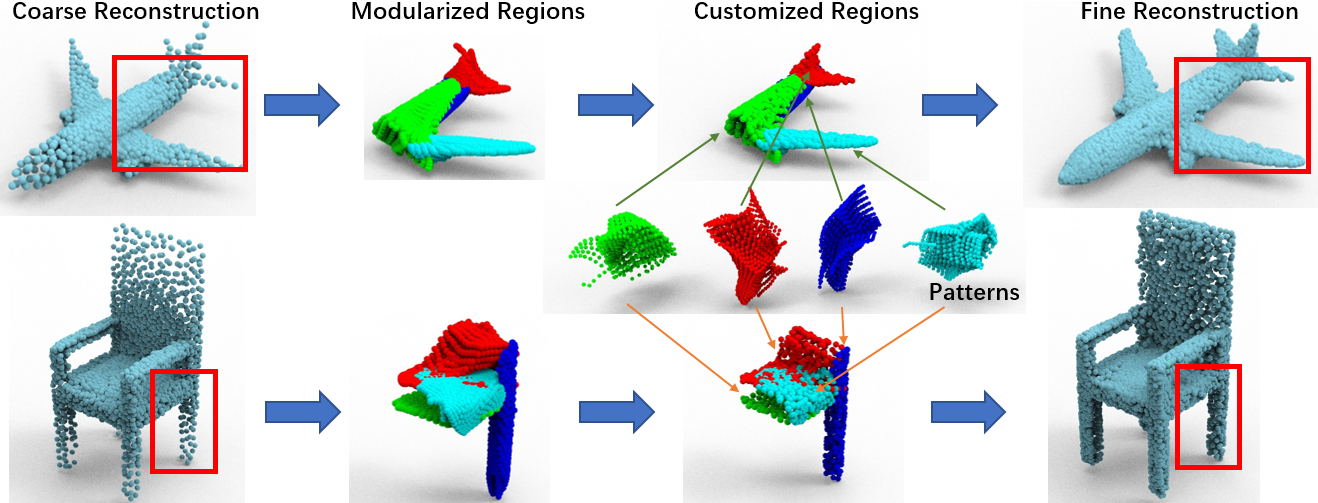}
  %
  %
  \vspace{-0.0in}
\caption{\label{fig:1}Overview Visualization.}
\vspace{-0.0in}
\end{figure}

\subsection{The Role of Pattern Learners }
Shapes from different classes may share some similar local structures, which makes it feasible to learn a local prior that is class-agnostic. Such as the part of the tail on a plane and the part of the back on a chair in Fig.~\ref{fig:1}. We aim to use pattern learners to learn a set of patterns that are class-agnoistic and shape-agnostic. These patterns are basic elements that can be assembled into a part without any bias on classes or shapes.

\subsection{Class-agnostic on initial prediction }
\begin{figure}[h]
\vspace{-0.0in}
  \centering
   \includegraphics[width=\linewidth]{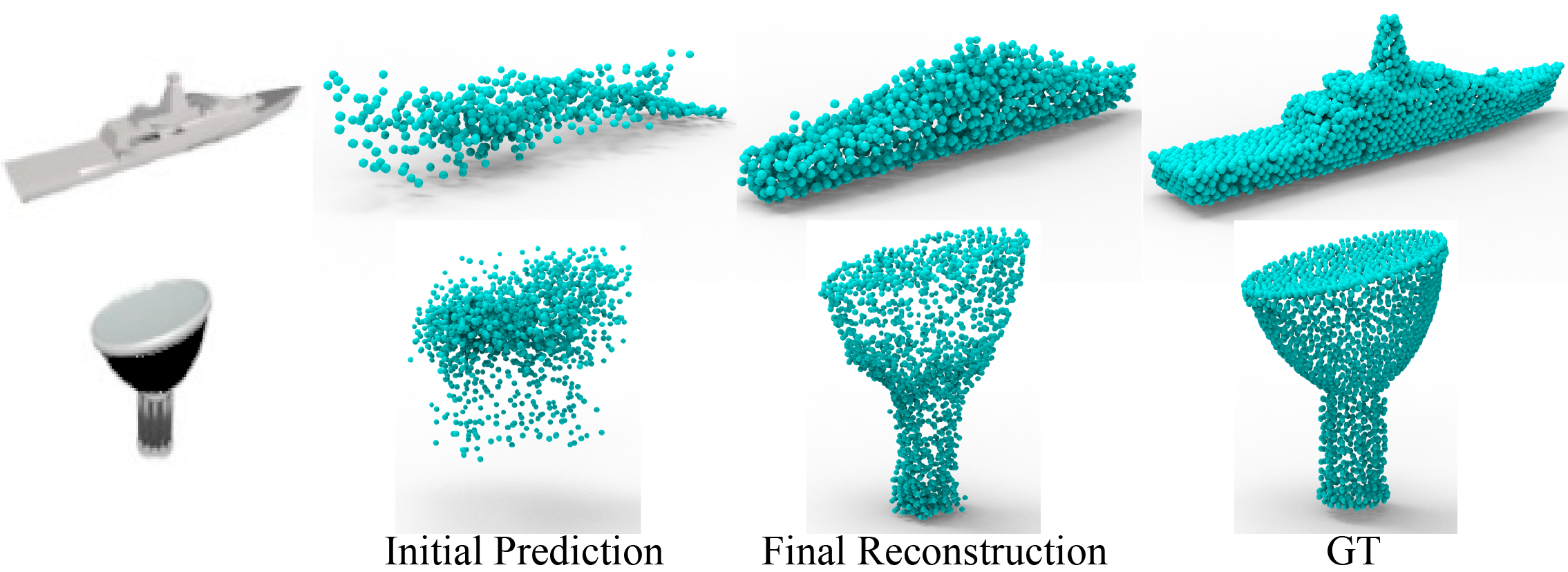}
  %
  %
  \vspace{-0.0in}
\caption{\label{fig:overview12}Shape reconstruction from unseen class.
}

\end{figure}

We want our initial prediction to keep the global prior weak and not biased to seen classes during training. This can be easily achieved by using a small weight to the loss supervising the initial prediction. In our experiment, we keep the loss weight $\alpha=0.1$ and produce a class-agnostic initial prediction. If we lower the loss weight, the initial prediction gets worse, but our final reconstruction is still plausible. This is because, in this case, pattern learners, modularizers and customizers must get used to starting from a bad initial prediction, and focus more on the image feature to learn our local prior. This leads to a class-agnostic prior that generalizes well and do not rely on the global prediction, even though the initial prediction may be bad. Fig.~\ref{fig:overview12} shows that we can reconstruct unseen objects well but merely from bad initial predictions.

\begin{table*}[h]
\vspace{-0.0in}
\centering
\resizebox{\linewidth}{!}{
    \begin{tabular}{c|c|c|c|c|c|c|c|c|c|c|c|c}  
     \hline
    \multicolumn{2}{c|}{Method}&Bench&Vessel&Rifle&Sofa&Table&Phone&Cabinet&Speaker&Lamp&Display&Mean\\
     \hline
     \multirow{4}{*}{Viewer-Centered}&DRC&0.120&0.109&0.121&0.107&0.129&0.132&0.142&0.141&0.131&0.156&0.129\\
     &MarrNet&0.107&0.094&0.125&0.090&0.122&0.117&0.125&0.123&0.144&0.149&0.120\\
     &Multi-View&0.092&0.092&0.102&0.085&0.105&0.110&0.119&0.117&0.142&0.142&0.111\\
     &GenRe&0.089&0.092&0.112&0.082&0.096&0.107&0.116&0.115&0.124&0.130&0.106\\
    \hline
    \hline
    \multirow{6}{*}{Object-Centered}&DRC&0.112&0.100&0.104&0.108&0.133&0.199&0.168&0.164&0.145&0.188&0.142\\
     &AtlasNet&0.102&0.092&0.088&0.098&0.130&0.146&0.149&0.158&0.131&0.173&0.127\\
     &GraphX&0.111&0.065&0.119&0.098&0.138&0.120&0.113&0.111&0.134&0.114&0.112\\
     &DRWR&0.075&0.059&0.104&0.070&0.100&0.094&0.088&0.086&0.102&0.097&0.088\\
     &CD&0.110&0.084&0.121&0.122&0.114&0.136&0.126&0.122&0.143&0.160&0.124\\
     \cline{2-13}
     &Ours&\textbf{0.054}&\textbf{0.046}&\textbf{0.046}&\textbf{0.058}&\textbf{0.070}&\textbf{0.061}&\textbf{0.071}&\textbf{0.072}&\textbf{0.089}&\textbf{0.077}&\textbf{0.064}\\
     \hline
   \end{tabular}}
   \vspace{-0.0in}
   \caption{Accuracy of reconstruction with $1024$ points for unseen classes under ShapeNet in terms of L1-CD.}  
   \label{table:NOX21}
   \vspace{-0.0in}
\end{table*}

\begin{figure}[h]
\vspace{-0.0in}
  \centering
   \includegraphics[width=0.9\linewidth]{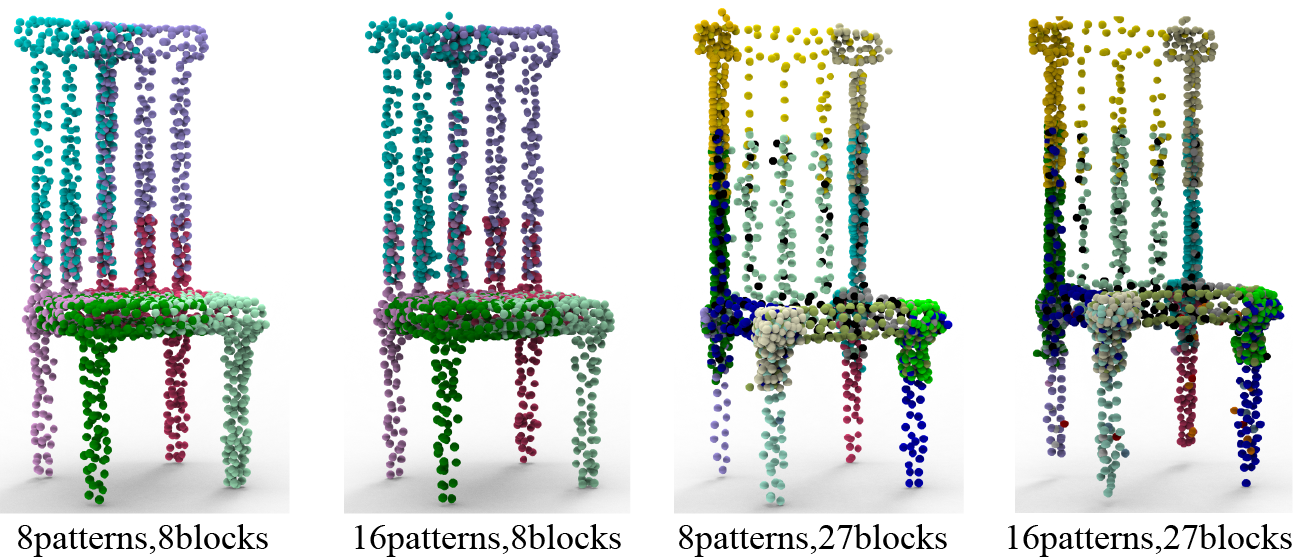}
  %
  %
  \vspace{-0.0in}
\caption{\label{fig:robust}Robustness to Parameters.
}
\vspace{-0.0in}
\end{figure}

\subsection{Robustness to Parameters }
We explore the effect of parameters. Here the parameters mainly include pattern number and region number. We compare the effect of different number of patterns and different number of regions on the final reconstruction. Fig.~\ref{fig:robust} shows that there is not too much difference with different parameters.

\subsection{Computation Complexity }
Our network is merely formed by a CNN and MLPs. It (31.51M) has even fewer parameters than DRWR~\cite{handrwr2020}: 35.94M, GraphX~\cite{DBLP:conf/iccv/NguyenCK019}: 38.87M. It (604.9s / 4000 iter) can also be trained faster (DRWR: 1630.5s, GraphX: 1637.4s).

%

\section{More Results}
\subsection{Reconstruction from Seen Classes }
\begin{figure*}[tb]
  \centering
   \includegraphics[width=\linewidth]{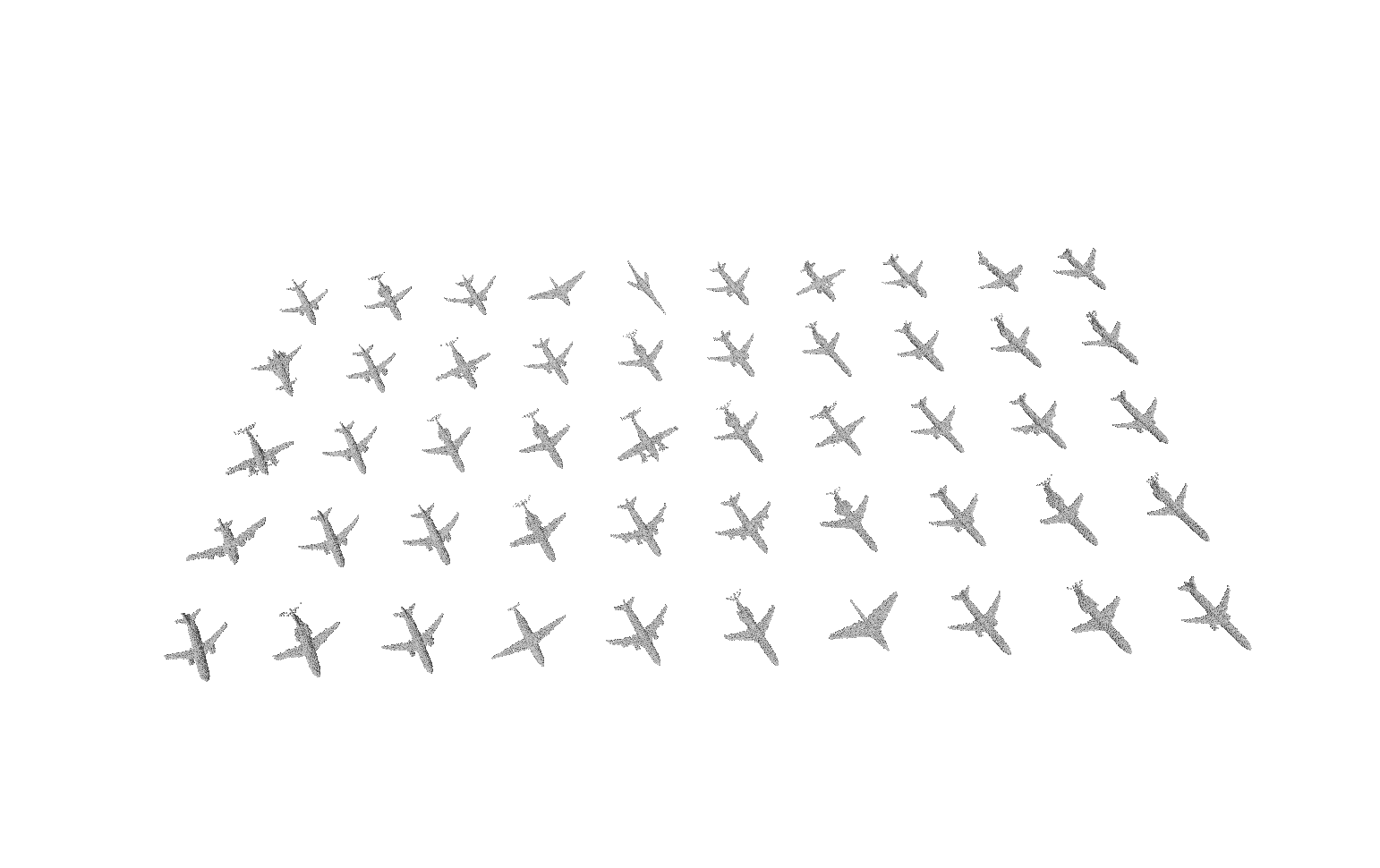}
  %
  %
\caption{\label{fig:Overview}Our point cloud reconstruction from seen classes. The 50 shapes are from Plane class under ShapeNet.
}
\end{figure*}

\begin{figure*}[tb]
  \centering
   \includegraphics[width=\linewidth]{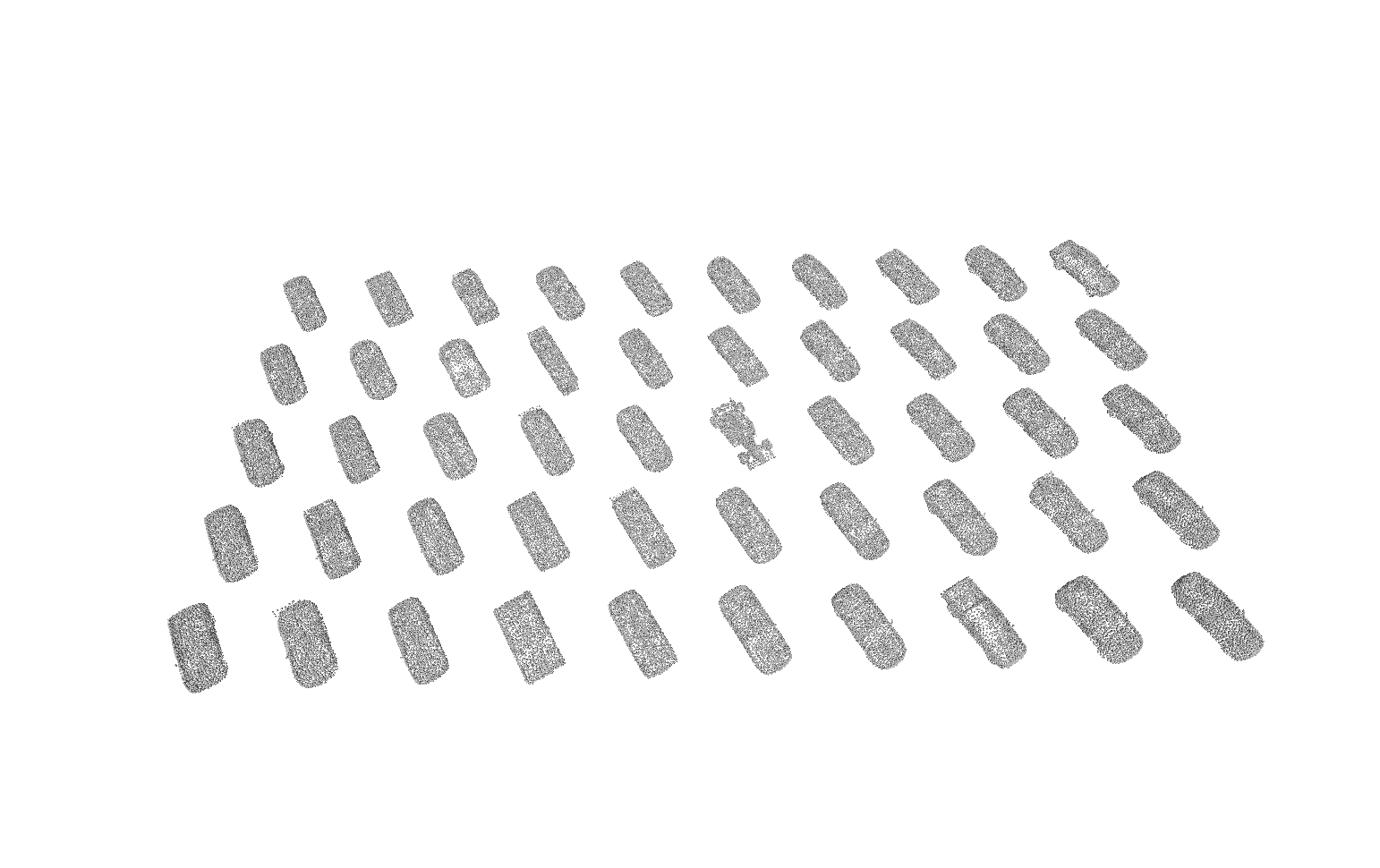}
  %
  %
\caption{\label{fig:Overview1}Our point cloud reconstruction from seen classes. The 50 shapes are from Car class under ShapeNet.
}
\end{figure*}

\begin{figure*}[tb]
  \centering
   \includegraphics[width=\linewidth]{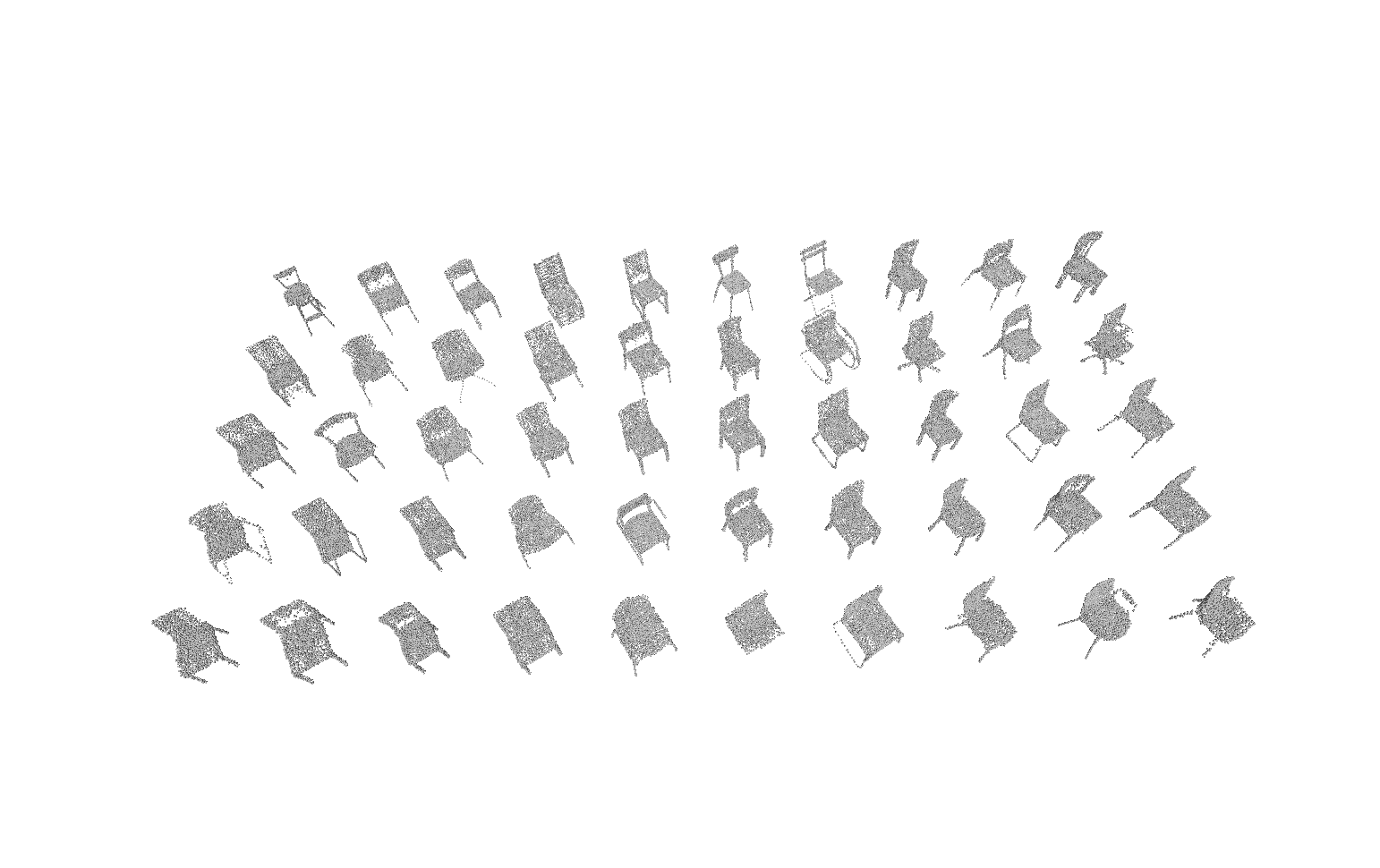}
  %
  %
\caption{\label{fig:Overview2}Our point cloud reconstruction from seen classes. The 50 shapes are from Chair class under ShapeNet.
}
\end{figure*}

We visualize more point clouds reconstructed from seen classes under ShapeNet dataset~\cite{ChangFGHHLSSSSX15}. We train our model under all classes in ShapeNet and test our model under the same set of seen classes. We show 50 reconstructed point clouds which are randomly sampled from each one of Plane, Car, Chair classes, respectively, in Fig.~\ref{fig:Overview}, Fig.~\ref{fig:Overview1}, and Fig.~\ref{fig:Overview2}. The high-fidelity reconstruction demonstrates that our method can reveal accurate geometry details.

\subsection{Reconstruction from Unseen Classes }
\begin{figure*}[tb]
  \centering
   \includegraphics[width=\linewidth]{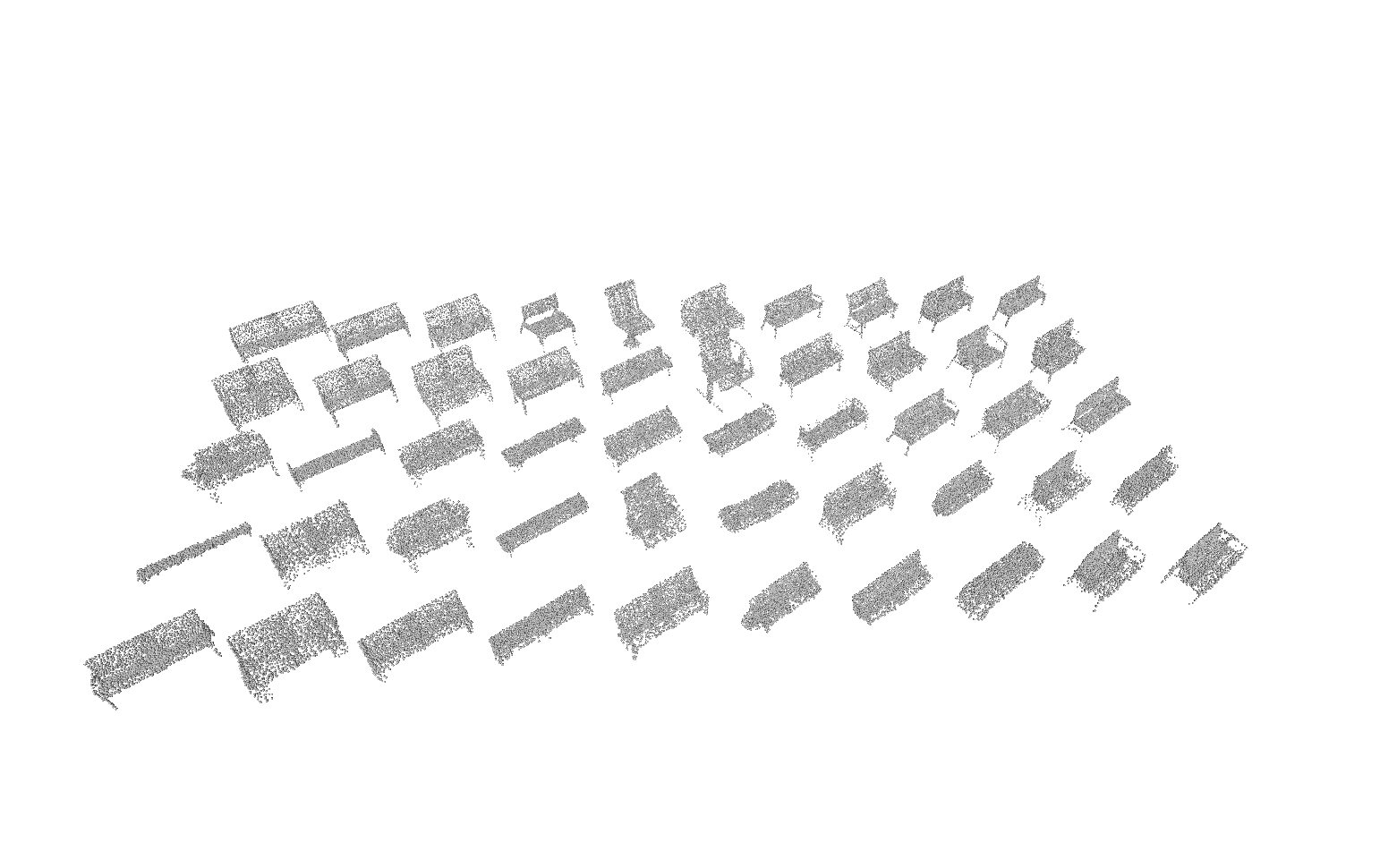}
  %
  %
\caption{\label{fig:Overview3}Our point cloud reconstruction from unseen classes. The 50 shapes are from Bench class under ShapeNet.
}
\end{figure*}

\begin{figure*}[tb]
  \centering
   \includegraphics[width=\linewidth]{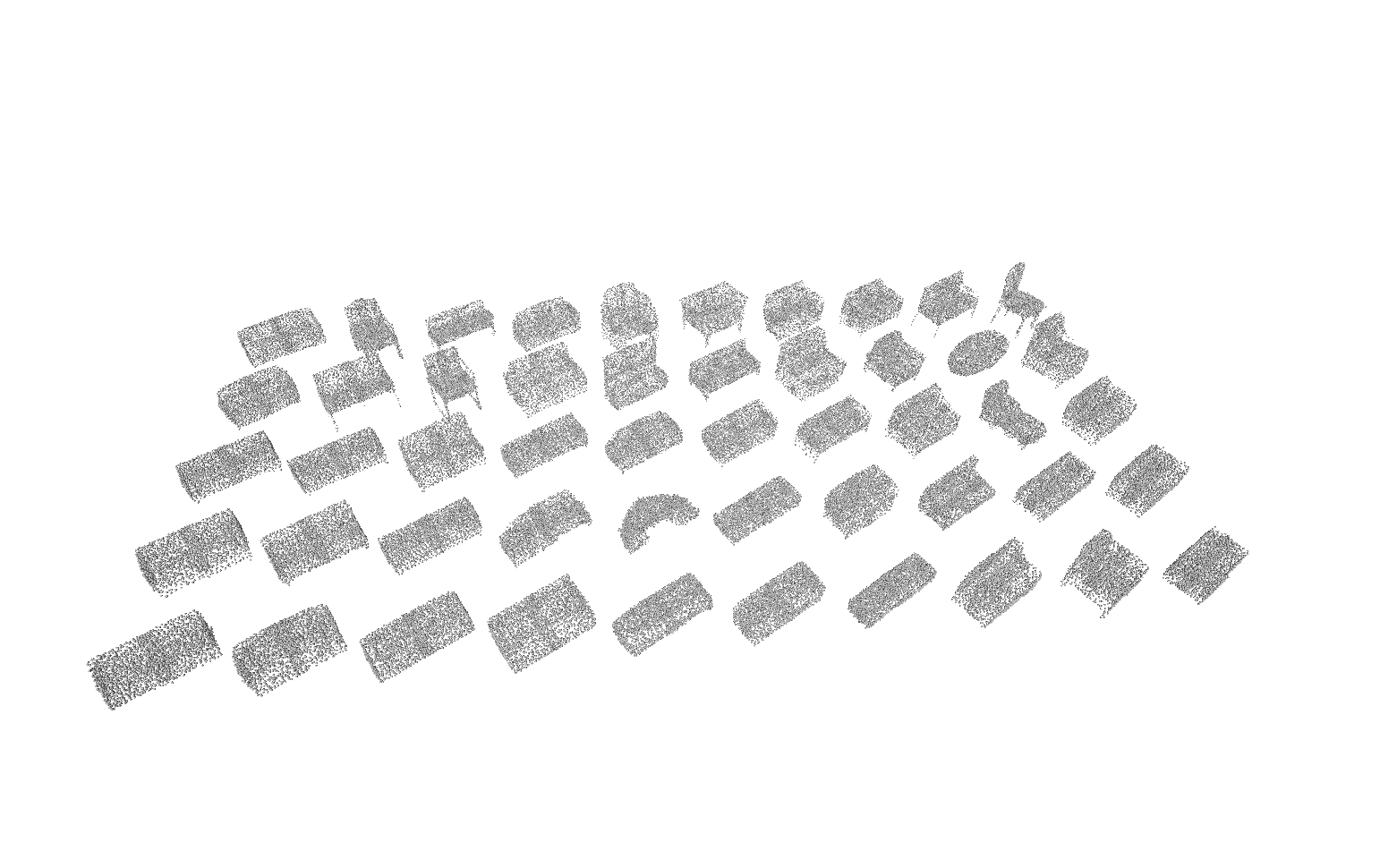}
  %
  %
\caption{\label{fig:Overview4}Our point cloud reconstruction from unseen classes. The 50 shapes are from Sofa class under ShapeNet.
}
\end{figure*}

\begin{figure*}[tb]
  \centering
   \includegraphics[width=\linewidth]{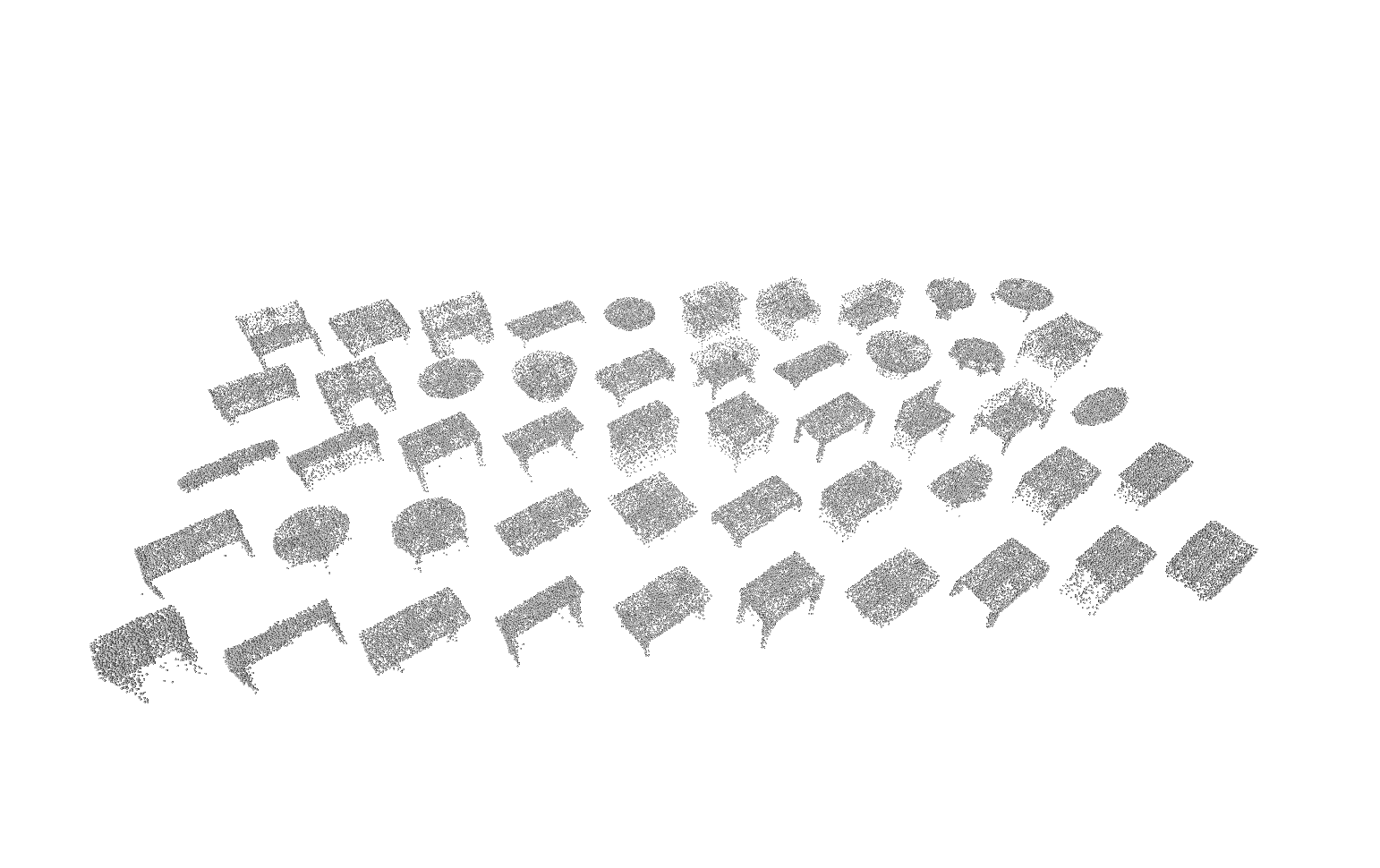}
  %
  %
\caption{\label{fig:Overview5}Our point cloud reconstruction from unseen classes. The 50 shapes are from Table class under ShapeNet.
}
\end{figure*}

We first report our numerical comparison under 10 unseen classes in ShapeNet in Table~\ref{table:NOX21}. The comparison shows that our method achieves higher accuracy over all classes than other methods. Then we visualize more point clouds reconstructed from unseen classes under ShapeNet dataset~\cite{ChangFGHHLSSSSX15}. We train our model under Chair, Car, and Plane, but test our model under 10 unseen classes under ShapeNet. We show 50 reconstructed point clouds from each one of the 3 classes including Bench, Sofa, and Table, in Fig.~\ref{fig:Overview3}, Fig.~\ref{fig:Overview4}, and Fig.~\ref{fig:Overview5}, respectively.

\begin{figure*}[tb]
  \centering
   \includegraphics[width=\linewidth]{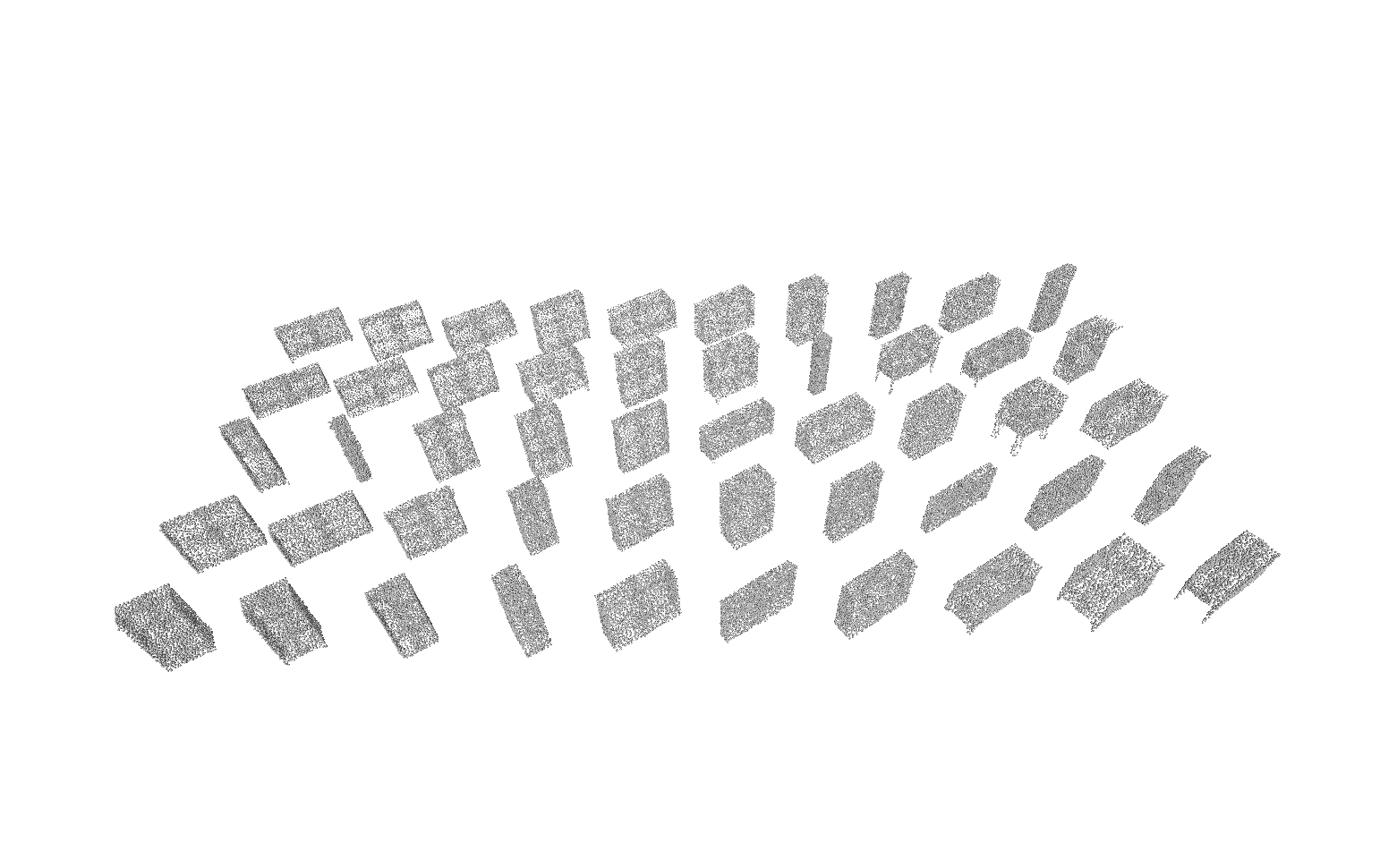}
  %
  %
\caption{\label{fig:Overview6}Our point cloud reconstruction from unseen classes. The 50 shapes are from Wardrobe class under Pixel3D.
}
\end{figure*}

\begin{figure*}[tb]
  \centering
   \includegraphics[width=\linewidth]{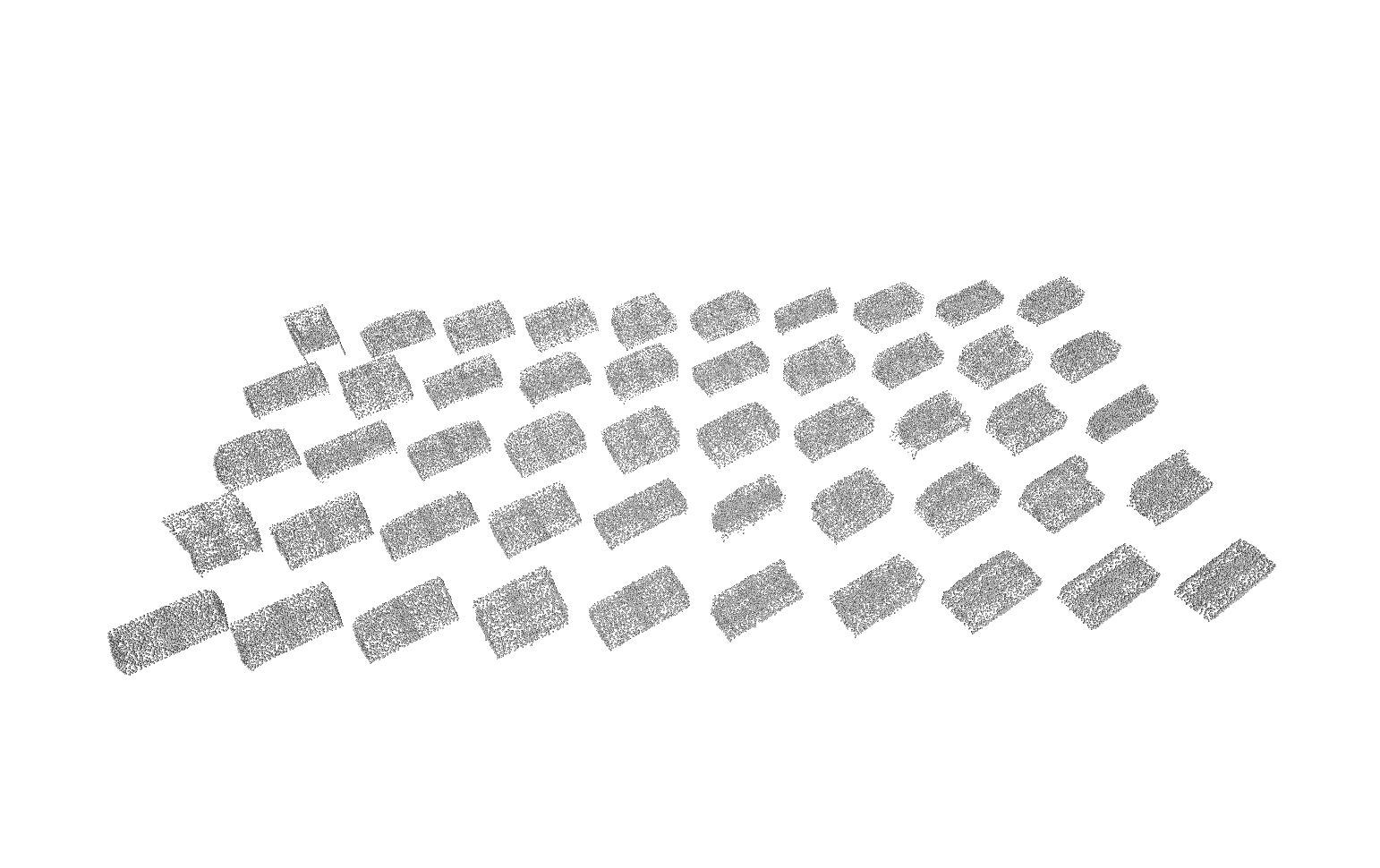}
  %
  %
\caption{\label{fig:Overview7}Our point cloud reconstruction from unseen classes. The 50 shapes are from Sofa class under Pixel3D.
}
\end{figure*}

\begin{figure*}[tb]
  \centering
   \includegraphics[width=\linewidth]{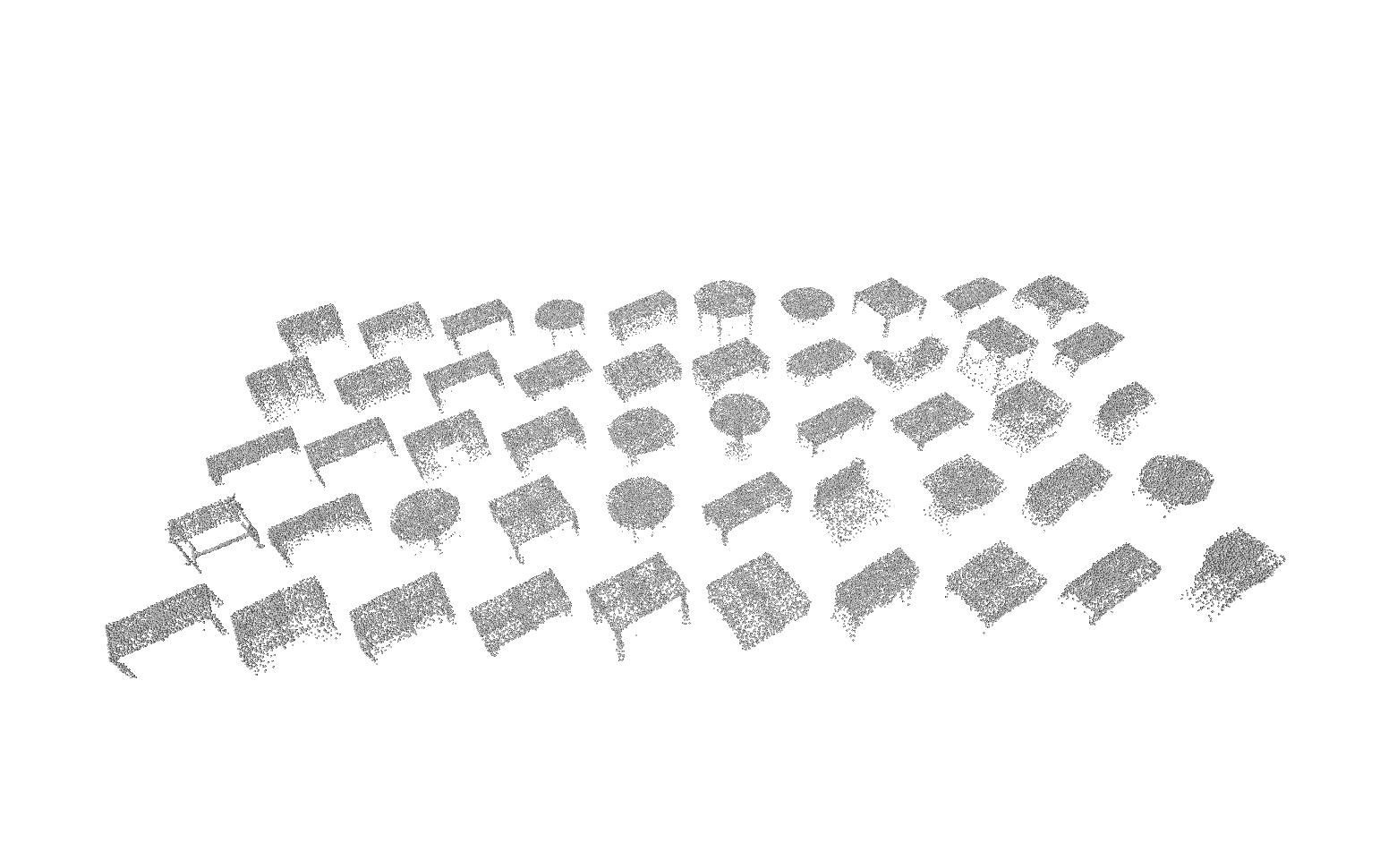}
  %
  %
\caption{\label{fig:Overview8}Our point cloud reconstruction from unseen classes. The 50 shapes are from Table class under Pixel3D.
}
\end{figure*}

Then, we visualize the point clouds reconstructed from unseen classes under Pixel3D~\cite{pix3d}, which is a large-scale real dataset. We use the same model trained under Chair, Car, and Plane in ShapeNet, but test the model under classes in Pixel3D including Bed, Bookcase, Desk, Sofa, and Wardrobe. We show 50 reconstructed point clouds from each one of the 3 classes including Wardrobe, Sofa, and Table, respectively, in Fig.~\ref{fig:Overview6}, Fig.~\ref{fig:Overview7}, and Fig.~\ref{fig:Overview8}.

These results demonstrate that our method can reconstruct point clouds in an object-centered coordinate system from unseen classes in high accuracy, which justifies that our local prior learned by customizing pattern modularized regions can generalize to unseen classes very well.

\section{Optimization Visualization}
We visualize the shape optimization in our video. Please watch our video for more details.

%
%
\bibliographystyle{splncs04}
\bibliography{papers}